\documentclass{article}

\usepackage{PRIMEarxiv}

\usepackage[utf8]{inputenc} 
\usepackage[T1]{fontenc}    
\usepackage{hyperref}       
\usepackage{url}            
\usepackage{booktabs}       
\usepackage{amsfonts, amsmath, amssymb}       
\usepackage{nicefrac}       
\usepackage{microtype}      
\usepackage{lipsum}
\usepackage{fancyhdr}       
\usepackage{graphicx}       
\usepackage{subcaption}
\usepackage{soul, xcolor}
\graphicspath{{media/}}     
\usepackage{mathabx}
\usepackage{amsmath}
\usepackage{mathrsfs}
\usepackage{mathtools}
\usepackage{multicol}
\usepackage{amssymb}
\usepackage{enumerate}
\usepackage{dsfont}
\usepackage{bm}
\usepackage{graphicx}
\usepackage{physics}
\usepackage{siunitx}
\usepackage{multicol}
\usepackage{xcolor}
\usepackage{centernot}
\usepackage{appendix}
\usepackage{authblk}
\usepackage{natbib}
\usepackage{tikz,tikz-cd}
\usepackage{soul}
\usepackage{float}
\usepackage{svg}
\usepackage{algorithm}
\usepackage{algpseudocode}
\usepackage{mathtools}
\usepackage{multirow}
\usepackage{caption}
\usepackage{subcaption}
\usepackage{cancel}
\usepackage{adjustbox}
\usepackage{rotating}
\usepackage{float}

\newcommand{\R}{\mathbb{R}}

\newcommand{\enc}{\boldsymbol{\phi}}
\newcommand{\dec}{\boldsymbol{\psi}}
\newcommand{\xirom}{\boldsymbol{\xi}}

\title{Deep Invertible Autoencoders for Dimensionality Reduction of Dynamical Systems}

\author[1,*]{\small Nicolò Botteghi}
\author[2]{\small Silke Glas}
\author[2]{\small Christoph Brune}

\affil[1]{\footnotesize MOX Laboratory, Department of Mathematics, Politecnico di Milano, Milano, Italy}
\affil[2]{\footnotesize Department of Applied Mathematics, University of Twente, Enschede, Netherlands}
\affil[*]{\footnotesize Corresponding author,  \texttt{nicolo.botteghi@polimi.it}}

\begin{document}
\maketitle
\begin{abstract}
Constructing reduced-order models (ROMs) capable of efficiently predicting the evolution of parameter-dependent high-dimensional dynamical systems is crucial in many applications in engineering and applied sciences. A popular class of projection-based ROMs projects the high-dimensional full-order model (FOM) dynamics onto a low-dimensional manifold. These projection-based ROMs approaches often rely on classical model reduction techniques such as proper orthogonal decomposition (POD) or, more recently, on neural network architectures such as autoencoders (AEs). In the case that the ROM is constructed by the POD, one has approximation guaranteed based based on the singular values of the problem at hand. However, POD-based techniques can suffer from slow decay of the singular values in transport- and advection-dominated problems. In contrast to that, AEs allow for better reduction capabilities than the POD, often with the first few modes, but at the price of theoretical considerations. In addition, it is often observed, that AEs exhibits a plateau of the projection error with the increment of the dimension of the trial manifold. \\
In this work, we propose a deep invertible AE architecture, named 
\emph{inv-AE}, that computationally improves upon the stagnation of the  reconstruction error typical of traditional AE architectures, e.g., convolutional, and the reconstructions quality.
Inv-AE is composed of several invertible neural network layers that allows for gradually recovering more information about the FOM solutions the more we increase the dimension of the reduced manifold. Through the application of inv-AE to a parametric 1-dimensional Burgers' equation, a parametric 2-dimensional fluid flow around an obstacle with variable geometry, and a parametric 3-dimensional Korteweg-de Vries, we show that (i) inv-AE mitigates the issue of the characteristic plateau of (convolutional and fully connected) AEs and (ii) inv-AE can be combined with popular autoencoder-based ROM approaches, e.g., DL-ROM and POD-DL-ROM, to improve their accuracy. 
\end{abstract}

\keywords{Model-order Reduction \and Deep Invertible Autoencoder \and Nonlinear Dimensionality Reduction}

\section{Introduction}

Many applications in engineering and applied science, e.g., fluid dynamics, chemical and process engineering, climate science, and (soft) robotics, require accurate physical modeling of the systems by means of partial differential equations (PDEs). Often, these systems do not yield closed-form solutions such that numerical simulations are needed to provide approximations. Usually, the solution of these high-dimensional models, which we refer to as full-order models (FOMs), is often computational demanding, limiting the application of such models in all the scenarios requiring real-time or iterative integration, e.g., in optimal control, reinforcement learning, or uncertainty quantification \cite{quarteroni2015reduced}. Reducing the computational cost of the FOMs by investigating new strategies for efficient model reduction (MOR) has been a research focus for decades and, more recently, the advances of machine learning had a significant influence on the development of data-driven reduced-order models (ROMs).

In case that the underlying equations are available, one popular option for building ROMs is by projection of the FOM equations onto a trial manifold, see, e.g.,  \cite{quarteroni2015reduced, benner2015survey, hesthaven2016certified, benner2017model, fresca2021comprehensive}. The trial manifold can be obtained using linear or nonlinear projection-based techniques such as proper orthogonal decomposition (POD)  or autoencoders (AE) \cite{goodfellow2016deep}. 
While linear projection techniques based on the singular value decomposition (SVD) have solid mathematical foundations and implementations, they may suffer from slow decay of the singular values in transport- and advection-dominated problems \cite{OhlR16}. The slow decay of the singular values negatively affects the reduction capabilities of the linear-subspace ROMs, making the dimension of the trial manifold extremely large, if high accuracy is required, and thus might not allow a drastic reduction of the computational cost of the ROM compared to the FOM. This is due to the fact, that linear-subspace ROMs have a lower bound for the approximation quality, the so-called Kolmogorov $n$-widths. The Kolmogorov $n$-widths denote the best-possible error for a linear-subspace projection-based ROM of size $n$, however, in most cases, there is no explicit construction available how this best-possible approximation error can be attained. 

The recent advances of machine learning have made nonlinear projection-based techniques, such as AEs, more and more appealing. AE-based methods can improve the reduction capabilities of linear-subspace MOR methods in challenging problems, showing better decay of the approximation errors, especially within the first few eigenfunctions \cite{champion2019data, maulik2021reduced, fukami2021sparse, fresca2021comprehensive, bakarji2023discovering, franco2024latent}. However, the use of neural networks (NNs) often complicated theoretical consideration of the approximation bounds of the techniques. Additionally, as observed in e.g., \cite{lee2020model,KimCWZ22,BucGH21}, the reconstruction error of AEs, namely the error due to the reduction of dimensionality, with respect to the dimension of the trial manifold, exhibits a plateau (often in the order of $10^{-3}$) after approximately reaching the intrinsic dimension of the underlying problem. This critical drawback of AE architecture poses a strict lower bound on the approximation capabilities of the resulting ROMs. 

AEs utilize two NNs to approximate the reconstruction mappings that are commonly referred to in the machine learning community as the \emph{encoder} and the \emph{decoder}. However, the encoder is (usually) a non-invertible function
that does not allow the decoder to properly recover the inputs accurately. The non-invertibility of the encoder makes the decoder process ambiguous and causes loss of information \cite{nguyen2019training}. 
Increasing the latent dimension does not mitigate this drawback and it explains the plateau observed in \cite{lee2020model} when AEs are used for building ROMs. Building symmetric AEs by construction has been shown by, e.g., \cite{otto2023learning, brivio2025deep} to be an effective strategy to improve upon traditional AE architectures in terms of performance and theoretical guarantees. However, the construction of such symmetric AEs comes at the price of relying on specialized cost functions or additional computational complexity.

The construction of \emph{analytically invertible} NNs layers, such as coupling layers, invertible residual networks, and Glow-type layers \cite{dinh2014nice, dinh2016density, gomez2017reversible, behrmann2019invertible, jacobsen2018revnet, kingma2018glow} has gained attention due to the numerous methods and applications that can benefit from them, e.g., generative models, density estimation, normalizing flows, generative adversarial \cite{rezende2015variational, grover2018flow, grathwohl2018ffjord, chen2019residual, ardizzone2019guided}, bi-directional training and efficient backpropagation \cite{gomez2017reversible}, inverse problems \cite{ardizzone2018analyzing, glaws2022invertible}, fundamental analysis \cite{behrmann2018analysis, gilmer2018adversarial, jacobsen2018excessive, behrmann2021understanding}, and, recently, dimensionality reduction with autoencoders \cite{nguyen2019training, teng2019invertible, li2021invertible}. However, it is worth mentioning that the analytical invertibility of invertible NNs do not necessarily translates into their \emph{numerical invertibility}. As highlighted in \cite{behrmann2021understanding}, invertible NNs often experience exploding inverses due to numerical instabilities and errors.  The Lipschitz constant of the invertible NNs determines how much these errors are magnified or attenuated. Therefore, controlling the Lipschitz constant is crucial for the numerical stability and invertibility of invertible NNs. Different invertible NN layers have different stability properties, but in general numerical errors become more severe the more we increase the depth of the models. Spectral normalization \cite{miyato2018spectral}, sigmoid scaling \cite{ardizzone2019guided}, and weight initialization \cite{kingma2016improved} can be used to control the local stability of the invertible NNs. However, global stability of the forward and backward pass is often hard to guarantee \cite{behrmann2021understanding}.

\begin{figure}[h!]
    \centering
    \includegraphics[width=0.95\linewidth]{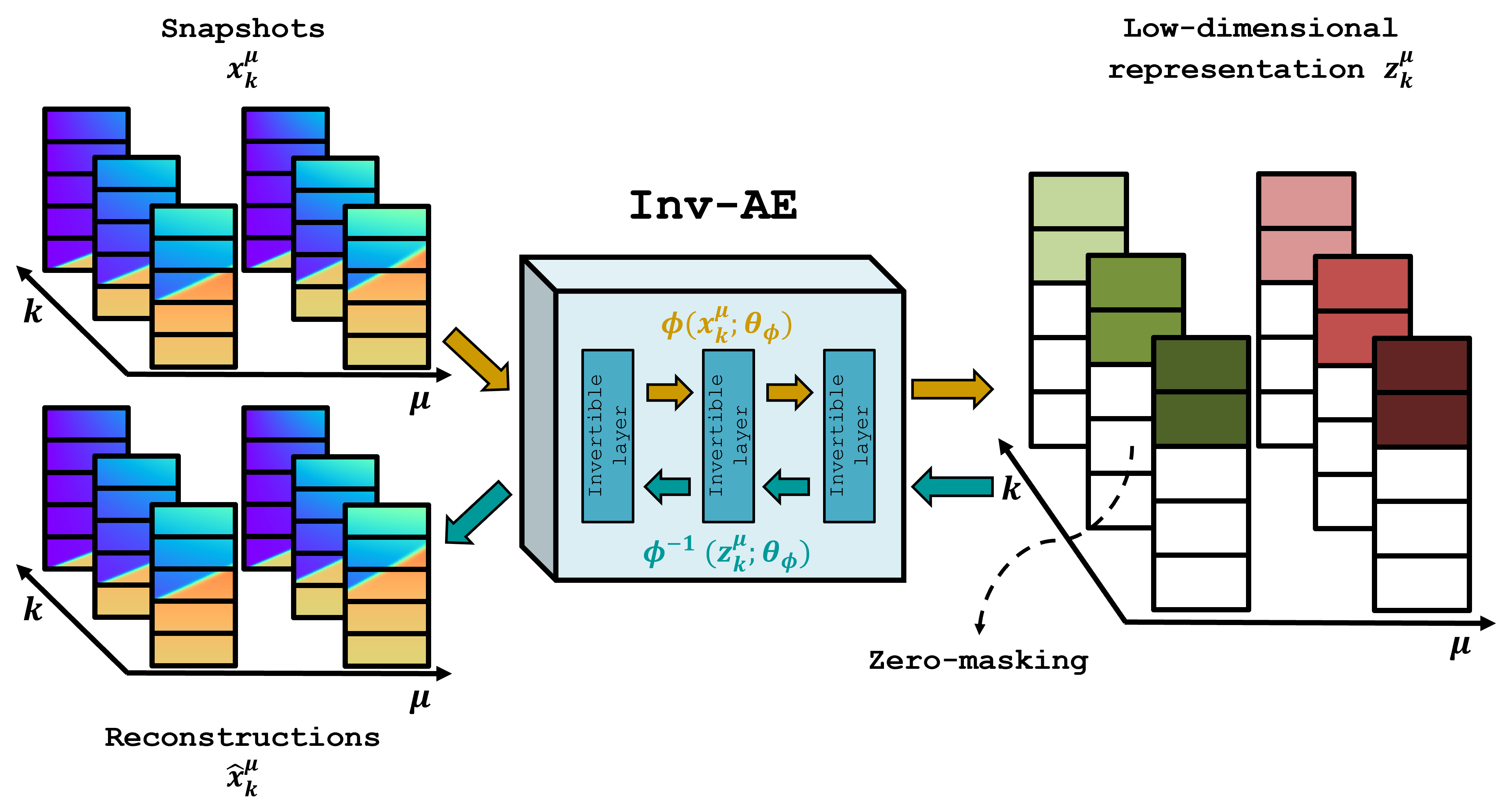}
    \caption{Dimensionality reduction with inv-AE. The snapshots $\bm{x}_k^{\boldsymbol{\mu}}$ are encoded using the encoder $\enc(\bm{x}^{\boldsymbol{\mu}}_k;\boldsymbol{\theta}_{\enc})$, successively zero-masked up to a chosen latent dimension, and eventually decoded using the inverse of the encoder $\enc^{-1}(\bm{z}^{\boldsymbol{\mu}}_k;\boldsymbol{\theta}_{\enc})$. }
    \label{fig:1a}
\end{figure}

In this work, we focus on enhancing nonlinear projection-based ROMs. In particular, we propose an analytically and numerically invertible AE (\emph{inv-AE}) architecture, tailored to MOR, that overcomes the reconstruction error plateau characterizing traditional AE architectures composed of feedforward or convolutional NN layers. To construct an inv-AE, we stack multiple \emph{invertible affine coupling layers} \cite{dinh2014nice, dinh2016density} to build an invertible NN. The forward pass through the invertible NN represents the encoder $\enc$, while the backward pass represents the decoder $\dec$ that in this case is equal to the inverse of the encoder by construction, i.e., $\dec=\enc ^{-1}$ (restricted to its image). In this way, through \emph{parameter sharing} of encoder and decoder, we are capable of drastically reducing the number of learnable parameters, namely weights and biases of the NN, and consequently the training cost.
With reference to Figure \ref{fig:1a}, a generic snapshot $\bm{x}^{\boldsymbol{\mu}}_k$, where $k$ denotes the time step and $\boldsymbol{\mu}$ the parametric dependency of the snapshot, is projected by the invertible layers, with learnable parameters $\boldsymbol{\theta}_{\enc}$, to $\enc(\bm{x}^{\boldsymbol{\mu}}_k;\boldsymbol{\theta}_{\enc})$. An invertible NN requires the inputs to be of the same dimension of its outputs. Thus, to reduce the dimensionality and promote the compression of information onto a lower-dimensional manifold, we zero-mask $\enc(\bm{x}^{\boldsymbol{\mu}}_k;\boldsymbol{\theta}_{\enc})$ up to the chosen latent dimensionality to obtain $\bm{z}^{\boldsymbol{\mu}}_k$. After zero-masking, $\bm{z}^{\boldsymbol{\mu}}_k$ is used fed to the inverse mapping $\enc^{-1}(\bm{z}^{\boldsymbol{\mu}}_k;\boldsymbol{\theta}_{\enc})$, i.e., the decoder of traditional AEs, to recover the reconstruction $\hat{\bm{x}}^{\boldsymbol{\mu}}_k$ of $\bm{x}^{\boldsymbol{\mu}}_k$. With this simple zero-masking, inspired by \cite{nguyen2019training}, we have transformed an invertible NN into an inv-AE. 

\begin{figure}[h!]
    \centering
    \includegraphics[width=0.95\linewidth]{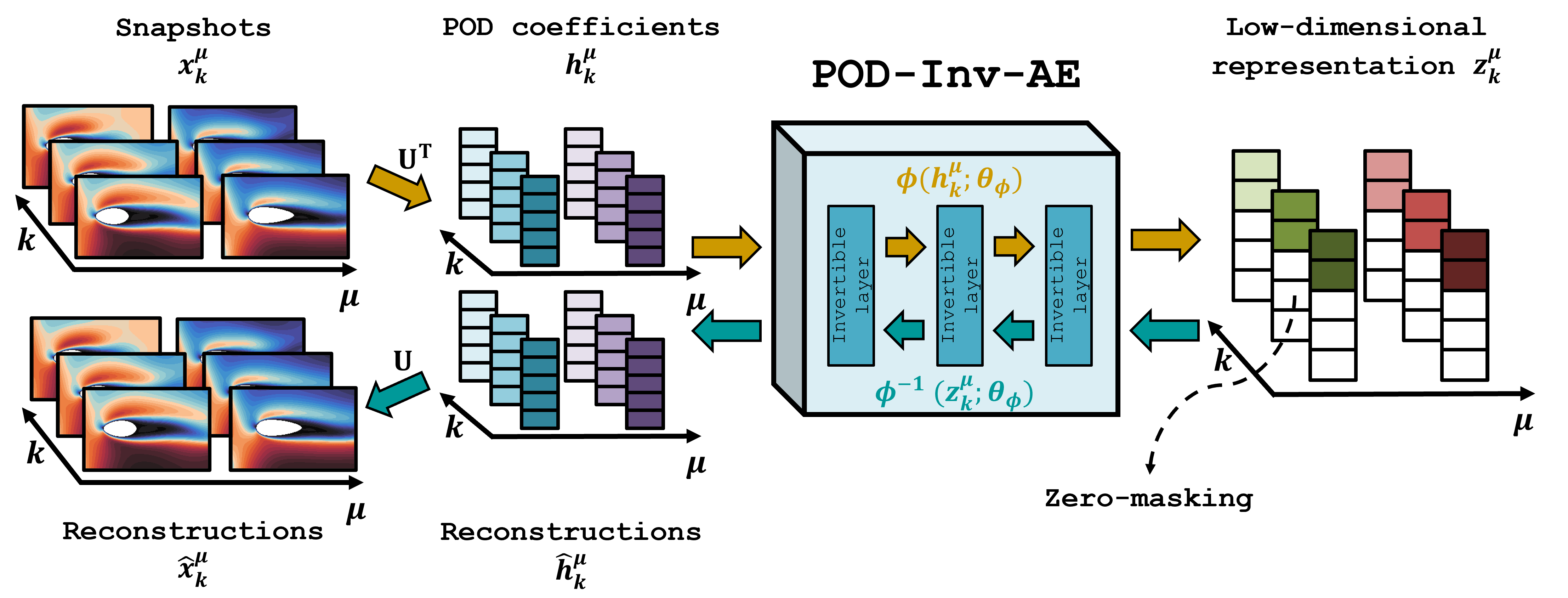}
    \caption{Dimensionality reduction with  POD-inv-AE. The snapshots dimensionality is first reduced using POD, then the POD coefficients are further compressed by zero-masking output of encoder $\enc(\bm{x}^{\boldsymbol{\mu}}_k;\boldsymbol{\theta}_{\enc})$. The POD coefficients are recovered by decoding the latent variables $\enc^{-1}(\bm{z}^{\boldsymbol{\mu}}_k;\boldsymbol{\theta}_{\enc})$, and eventually the snapshots are reconstructed from the POD coefficients using $U$.}
    \label{fig:1b}
\end{figure}

In many complex problems, such as in fluid dynamics, the snapshots can be extremely high-dimensional. However, it is possible to first perform dimensionality-reduction using SVD on the data matrix
\begin{equation*}
    \bm{X}_{\text{train}} \approx \bm{U} \boldsymbol{\Sigma} \bm{V}^{\top}\, ,
    \label{eq:SVD}
\end{equation*}
where $\bm{U} \in \mathbb{R}^{N\times r}$, $\boldsymbol{\Sigma} \in \mathbb{R}^{r \times r}$, $\bm{V} \in \mathbb{R}^{(N_t \times |\mathcal{P}_{\text{train}}|)\times r}$,  with $r$ being the chosen number of modes used to represent the data. Using $\bm{U}^{\top}$ we can obtain the POD coefficient $\bm{h}_k^{\boldsymbol{\mu}}$ from $\bm{x}_k^{\boldsymbol{\mu}}$ and rely on the inv-AE to recover the POD coefficients $\bm{h}_k^{\boldsymbol{\mu}}$. Eventually, using $\bm{U}$ we can recover the reconstruction $\hat{\bm{x}}_k^{\boldsymbol{\mu}}$ of $\bm{x}_k^{\boldsymbol{\mu}}$ from the reconstructed POD coefficients $\hat{\bm{h}}_k^{\boldsymbol{\mu}}$. We refer to this approach as \emph{POD-inv-AE} and we depict it in Figure \ref{fig:1b}.

We numerically study the reduction capabilities of inv-AE and POD-inv-AE on a parametric Burgers' equation, analogously to \cite{lee2020model}, on a parametric fluid flow around a obstacle with variable geometry, and a parametric Korteweg-de Vries equation. Eventually, we show that inv-AE and POD-inv-AE can replace the traditional AE architectures of popular deep learning-based ROM techniques such as DL-ROM \cite{fresca2021comprehensive} and POD-DL-ROM \cite{fresca2022pod} to improve the accuracy of the predictions and consequently the performance of the resulting ROMs. 

The rest of the paper is organized as follows: Section \ref{sec:preliminaries} introduces problem settings, linear and nonlinear strategies for dimensionality reduction, and deep learning-based ROM. Section \ref{sec:method} presents the proposed invertible AE architecture and its combination with deep learning-based ROM. Section \ref{sec:results} shows the numerical experiments, results, and discusses the findings. Eventually, Section \ref{sec:conclusion} concludes the paper.

\section{Preliminaries}\label{sec:preliminaries}

In this section, we summarize existing background knowledge needed for further comprehension of the paper. We begin by stating the parametric initial value problem, its MOR, the limitations of linear-subspace MOR, and AEs in Section \ref{Sec:problem_setting}. Further, in Section \ref{sec:dl-rom} we introduce (POD) DL-ROMs based on AEs.

\subsection{Problem Setting and Limitations of Linear-subspace Model Reduction} \label{Sec:problem_setting}

We consider parametric initial value problems as our FOM model: for given $\bm{\mu} \in \mathcal{P} \subset \R^p, p \in \mathbb{N}$ and initial condition $\bm{x}_0(\boldsymbol{\mu})\in \R^{N}$, we seek the solution $\bm{x}(\cdot;\mu) \in C^1(\mathcal{I};\R^{N})$ such that
\begin{equation}
    \frac{\rm d}{\rm{d}t}\bm{x}(t;\boldsymbol{\mu}) = \bm{f}(t, \bm{x}(t;\boldsymbol{\mu});\boldsymbol{\mu}),  \ \ \ \ \ \bm{x}(t_0;\boldsymbol{\mu}) = \bm{x}_0(\boldsymbol{\mu}) \in \mathbb{R}^N,
    \label{eq:FOM}
\end{equation}
using time interval $\mathcal{I}=(t_0, t_f]$ and $t_0 < t_f < \infty$, and right-hand side $\bm{f}:\mathcal{I} \times \mathbb{R}^N \times \mathcal{P} \rightarrow \mathbb{R}^N$. By solving the FOM with a numerical integration scheme in time, we obtain a collection of snapshots of the type $\bm{x}(t;\boldsymbol{\mu}) \rightarrow \bm{x}(t_k;\boldsymbol{\mu})=\bm{x}^{\boldsymbol{\mu}}_k$ for $k=1, \cdots,N_t$ for each parameter $\boldsymbol{\mu}$ in a discrete training set $\mathcal{P}_{\text{train}} \subseteq \mathcal{P}$, where we set $N_{\textnormal{train}}:= |\mathcal{P}_{\text{train}} |$. The collection of snapshots compose our data matrix such that we arrive at
\begin{equation*}
  \bm{X}_{\text{train}} =\left[ \begin{array}{ccccc} | & | & \cdots & | & | \\
         {\bm{x}_{1}}^{\boldsymbol{\mu}_1} & {\bm{x}}_{2}^{\boldsymbol{\mu}_1} & \cdots & {\bm{x}}_{N_t - 1}^{\boldsymbol{\mu}_{N_{\text{train}}}}  &  {\bm{x}}_{N_t}^{\boldsymbol{\mu}_{N_{\textnormal{train}}}} \\
         | & | & \cdots & | & |   \end{array} \right] \in \mathbb{R}^{N \times (N_t \times N_{\text{train}})} \, .
\label{eq:data_matrix}
\end{equation*}

Given the FOM in \eqref{eq:FOM}, a ROM can be constructed via reconstruction with the two mappings previously denoted as encoder $\enc:\mathbb{R}^N \rightarrow \mathbb{R}^n$ and decoder $\dec:\mathbb{R}^n \rightarrow \mathbb{R}^N$ such that $\enc \circ \dec = \text{id}_{\R^n}$ and $n\ll N$. 
Two terms commonly used in MOR are the computational intense offline phase, which is dependent on the FOM dimension $N$ and in which the ROM is built, as well as the cost-effective online phase, where the ROM (ideally only depending on the latent dimension $n$) is evaluated.
Linear-subspace MOR methods employs linear mappings for $\enc$ and $\dec$, while in the case of MOR on manifolds, the mappings $\enc$ and $\dec$ are nonlinear functions.\footnote{For a general differential geometric formulation of MOR on manifolds, we refer the reader to \cite{BucGHU24}.} 

In the case, that the mapping $\enc, \dec$ are linear the quality of the resulting ROM approximation can be bounded from below by the Kolmogorov $n$-widths. In many applications, such as in elliptic and parabolic partial differential equations, see, e.g.,  \cite{Maday2002,Haasdonk13}, it is observed, that Kolmogorov $n$-widths can obtain an exponential decay. However, in the case of transport-dominated problems, the Kolmogorov $n$-widths may decay slowly \cite{OhlR16}. One possibility on how to break this Kolmogorov barrier is to employ nonlinear mapping for $\enc, \dec$ such as autoencoders.

AEs \cite{hinton1993autoencoders, goodfellow2016deep} are special types of NN architectures used for dimensionality reduction. An AE (i) maps the high-dimensional data onto a low-dimensional latent variable representation using the encoder and (ii) maps the low-dimensional latent variables back to the original high-dimensional data space using the decoder, i.e, $\bm{x} \approx \hat{\bm{x}} = \dec(\enc(\bm{x};\boldsymbol{\theta}_{\enc});\boldsymbol{\theta}_{\dec})$.
We denote the learnable parameters, namely weights and biases, of the encoder and decoder by $\boldsymbol{\theta}_{\enc}$ and $\boldsymbol{\theta}_{\dec}$, respectively. These learnable parameters $\boldsymbol{\theta}_{\enc}, \boldsymbol{\theta}_{\dec}$ of the AE are jointly optimized by minimizing the reconstruction loss over a (train) set $X$ on $M$ training samples that we indicate as $\{\bm{x}^{(\text{1})},\cdots, \bm{x}^{(\text{M})}\}$, i.e., 
\begin{equation}
\begin{split}
      \mathcal{L}(\boldsymbol{\theta}_{\enc}, \boldsymbol{\theta}_{\dec}) &= \frac{1}{M}\sum_{i=1}^M||\bm{x}^{(i)} - \hat{\bm{x}}^{(i)}||^2_2  
      = \frac{1}{M}\sum_{i=1}^M ||\bm{x}^{(i)} - \dec(\enc(\bm{x}^{(i)}; \boldsymbol{\theta}_{\enc}); \boldsymbol{\theta}_{\dec}) ||^2_2\, ,
\end{split}
\label{eq:recon_loss}
\end{equation}
where $\bm{x}^{(\text{i})}$ indicates the $i^{\text{th}}$ training data and $\hat{\bm{x}}^{(\text{i})}$ its reconstruction using the AE.
The dimension of the latent space is a hyper-parameter of the AE model that can be chosen using prior knowledge or through a trail-and-error approach. Moreover, for $\dec, \enc$ being nonlinear mappings, adjusted variants of limiting widths can be investigated, where we refer the interested reader to \cite{CohFMS23} for an overview of nonlinear widths.

\subsection{(POD) Deep Learning-based ROMs}\label{sec:dl-rom}

The method of Deep Learning-based ROM (DL-ROM) \cite{fresca2021comprehensive} is a popular and effective approach for learning ROMs from data. In this context, the generic $i^{\text{th}}$ data $\bm{x}^{(\text{i})}$ corresponds to the $i^{\text{th}}$ FOM snapshot $\bm{x}^{\boldsymbol{\mu}}_k$ that might have temporal and parametric dependencies. DL-ROM is composed of a (convolutional) encoder $\enc(\bm{x}^{\boldsymbol{\mu}}_k;\boldsymbol{\theta}_{\enc})$
and a (convolutional) decoder $\dec(\bm{z}^{\boldsymbol{\mu}}_k;\boldsymbol{\theta}_{\dec})$ 
that learn a low-dimensional representation $\bm{z}^{\boldsymbol{\mu}}_k$ of the snapshots and their reconstructions $\hat{\bm{x}}^{\boldsymbol{\mu}}_k$ in parallel. To efficiently predict the system's dynamics, DL-ROM uses a feedforward NN $\xirom(k, \boldsymbol{\mu};\boldsymbol{\theta}_{\xirom})$, with learnable parameters $\boldsymbol{\theta}_{\xi}$, to predict the latent vector $\tilde{\bm{z}}^{\boldsymbol{\mu}}_k$ from the timestep $t$ and parameter vector $\boldsymbol{\mu}$. We can then retrieve the approximated FOM snapshots by decoding $\tilde{\bm{z}}^{\boldsymbol{\mu}}_k$.  
We can optimize the DL-ROM parameters, namely $\boldsymbol{\theta}_{\enc}$, $\boldsymbol{\theta}_{\dec}$, and $\boldsymbol{\theta}_{\xirom}$, by minimizing over the loss function
\begin{equation}
\begin{split}
      \mathcal{L}_{\text{DL-ROM}}(\boldsymbol{\theta}_{\enc}, \boldsymbol{\theta}_{\dec}, \boldsymbol{\theta}_{\xirom})&=\frac{1}{M}\sum_{i=1}^M\Big(\frac{\alpha}{2}||\bm{x}^{(i)} - \tilde{\bm{x}}^{(i)}||^2_2 + \frac{1-\alpha}{2}||\bm{z}^{(i)} - \tilde{\bm{z}}^{(i)}||^2_2 \Big) \\
      &=\frac{1}{M}\sum_{i=1}^M\Big(\frac{\alpha}{2}||\bm{x}^{(i)} - \dec(\tilde{\bm{z}}^{(i)};\boldsymbol{\theta}_{\dec})||^2_2 + \frac{1-\alpha}{2}||\enc(\bm{x}^{(i)};\boldsymbol{\theta}_{\enc}) - \tilde{\bm{z}}^{(i)}||^2_2 \Big) \\
      &=\frac{1}{M}\sum_{i=1}^M\Big(\frac{\alpha}{2}||\bm{x}^{(i)} - \dec(\xirom(k^{(i)},\boldsymbol{\mu}^{(i)};\boldsymbol{\theta}_{\xirom});\boldsymbol{\theta}_{\dec})||^2_2 + \frac{1-\alpha}{2}||\enc(\bm{x}^{(i)};\boldsymbol{\theta}_{\enc}) - \xirom(k^{(i)},\boldsymbol{\mu}^{(i)};\boldsymbol{\theta}_{\xirom})||^2_2 \Big)\, , 
\end{split}
\label{eq:DL-ROM_training}
\end{equation}
where $0 \le \alpha \le 1$.
In the online phase, we use $\xirom(k,\boldsymbol{\mu};\boldsymbol{\theta}_{\xirom})$ to predict the evolution of the system at an arbitrary timestep $k$ for a given $\boldsymbol{\mu}$ and decode the latent vectors $\tilde{\bm{z}}^{\boldsymbol{\mu}}_k$ using the decoder to retrieve approximations to the FOM solutions. 

The POD-DL-ROM method \cite{fresca2022pod} extends DL-ROM with an additional POD reduction step before the AE neural network to drastically reducing the number of learnable parameters of the AE and consequently reduce the training times. First the POD is computed for a chosen dimension $r$, $n \le r \le N$. Using the POD basis $\bm{U} \in \R^{N \times r}$ the respective POD coefficients $\bm{h}^{\boldsymbol{\mu}}_k:=\bm{U}^{\top}\bm{x}^{\boldsymbol{\mu}}_k$ on the intermediate linear subspace $\R^{r}$ are computed. Subsequently, the DL-ROM is trained using \eqref{eq:DL-ROM_training} by replacing $\bm{x}^{\boldsymbol{\mu}}_k$ with the POD coefficients $\bm{h}^{\boldsymbol{\mu}}_k$. In the online phase, we follow the DL-ROM steps and afterwards recover the FOM solutions from the reconstructed POD coefficients, i.e., $\tilde{\bm{x}}^{\boldsymbol{\mu}}_k=\bm{U}\tilde{\bm{h}}^{\boldsymbol{\mu}}_k=\bm{U}\dec(\xirom(k,\boldsymbol{\mu};\boldsymbol{\theta}_{\xirom});\boldsymbol{\theta}_{\dec})$. Note, that for the POD-DL-ROM, an adjusted version of the Kolmogorov $n$-widths holds, due to the linear reduction step introduced by the POD, see \cite{BucGH24}.

\section{Invertible Autoencoders for Dimensionality and Model Reduction}\label{sec:method}

Traditional AE architectures utilize (deep) NNs to parametrize the encoder and decoder and minimize the reconstruction loss \eqref{eq:recon_loss} to optimize the learnable parameters of the two architectures. However, the encoder is usually non-invertible and, therefore, it cannot be inverted by the decoder, resulting in a loss of information \cite{nguyen2019training}. The loss of information is typically not reduced by increasing the dimension of the manifold, i.e., using more latent variables, and it is correlated with the reconstruction-error plateau firstly observed by \cite{lee2020model} in the context of data-driven ROM. To tackle the problem of (non-)invertibility of AE architectures, inspired by \cite{nguyen2019training}, we propose an \emph{analytically and numerically invertible} AE (\emph{inv-AE}) architecture capable of going beyond the reconstruction-error plateau of traditional AEs.

The rest of the section is organized as follows: Section \ref{subsec:inn} presents the building blocks of inv-AE, namely the invertible NN layers, while Section \ref{subsec:fromINNtoINVAE} presents the inv-AEs.  Eventually, Section \ref{subsec:invDLROM} introduces a new class of DL-ROMs named invertible (POD) DL-ROMs.

\subsection{Invertible Neural Networks and Layers}\label{subsec:inn}
Similarly to invertible NNs, the inv-AE is composed of multiple invertible layers. Inv-AE uses invertible affine coupling layers \cite{dinh2014nice, dinh2016density}, but other invertible layers can be simply plugged-in without changing the core principles. Each affine coupling layer is composed of four nonlinear learnable functions, scaling and translating the input $\bm{x}$: 
\begin{enumerate}
    \item $\bm{s}_1:\mathbb{R}^{N/2} \rightarrow \mathbb{R}^{N/2}$,
    \item $\bm{s}_2:\mathbb{R}^{N/2} \rightarrow \mathbb{R}^{N/2}$,
    \item $\bm{t}_1:\mathbb{R}^{N/2} \rightarrow \mathbb{R}^{N/2}$, and
    \item $\bm{t}_2:\mathbb{R}^{N/2} \rightarrow \mathbb{R}^{N/2}$\, ,
\end{enumerate}
where $N$ indicates the dimension of the input $\bm{x}$.
Each function is parametrized by NN with learnable parameter vectors $\boldsymbol{\theta}_{\bm{s}_1}$, $\boldsymbol{\theta}_{\bm{s}_2}$, $\boldsymbol{\theta}_{\bm{t}_1}$, $\boldsymbol{\theta}_{\bm{t}_2}$, respectively.
\begin{figure}[h!]
    \centering
    \includegraphics[width=\textwidth]{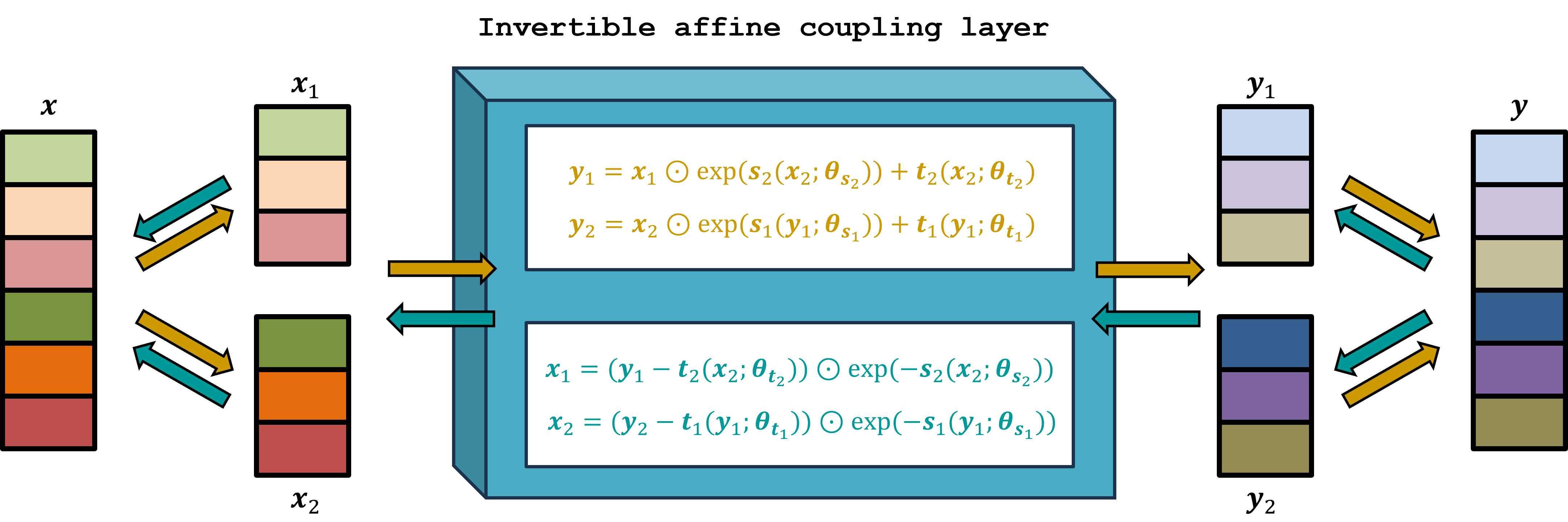}
    \caption{Invertible affine coupling layer. The forward pass through the layer is indicated by dark khaki arrows, while the backward pass by teal arrows.}
    \label{fig:2}
\end{figure}
With reference to Figure \ref{fig:2}, an affine coupling layer first splits the input $\bm{x}$ of dimension $N$ in two part of equal dimension, i.e., $\bm{x}_1, \bm{x}_2 \in \R^{N/2}$ such that $\bm{x}= \left( \bm{x}_1,\bm{x}_2 \right)^{\top}$. Then, the transformations $\bm{s}_1(\cdot;\boldsymbol{\theta}_{\bm{s}_1})$, $\bm{s}_2(\cdot;\boldsymbol{\theta}_{\bm{s}_2})$, $\bm{t}_1(\cdot;\boldsymbol{\theta}_{\bm{t}_1})$, and $\bm{t}_2(\cdot;\boldsymbol{\theta}_{\bm{t}_2})$ are applied to $\bm{x}_1$ and $\bm{x}_2$ to obtain the outputs $\bm{y}_1 \in \R^{N/2}$ and $\bm{y}_2 \in \R^{N/2}$ of the layer. The forward pass through an invertible coupling layer can be written by
\begin{equation*}
    \begin{split}
        \bm{y}_1 &= \bm{x}_1 \odot \exp(\bm{s}_2(\bm{x}_2);\boldsymbol{\theta}_{\bm{s}_2}) + \bm{t}_2(\bm{x}_2;\boldsymbol{\theta}_{\bm{t}_2})\\
        \bm{y}_2 &= \bm{x}_2 \odot \exp(\bm{s}_1(\bm{y}_1);\boldsymbol{\theta}_{\bm{s}_1}) + \bm{t}_1(\bm{y}_1; \boldsymbol{\theta}_{\bm{t}_1}) \\
    \end{split}\, ,
\label{eq:affine_layer_forward}
\end{equation*}
where $\odot$ indicates the Hadamard product and the output $\bm{y}$ is obtained by concatenation of $\bm{y}_1$, $\bm{y}_2$, i.e., $\bm{y}= \left( \bm{y}_1,\bm{y}_2 \right)^{\top} \in \R^{N}$. The coupling layer is analytically invertible, and we can write its backward pass from $\bm{y}$ to $\bm{x}$ by
\begin{equation*}
    \begin{split}
        \bm{x}_1 &= (\bm{y}_1 - \bm{t}_2(\bm{x}_2;\boldsymbol{\theta}_{\bm{t}_2})) \odot \exp(-\bm{s}_2(\bm{x}_2;\boldsymbol{\theta}_{\bm{s}_2})) \, , \\
        \bm{x}_2 &= (\bm{y}_2 - \bm{t}_1(\bm{y}_1;\boldsymbol{\theta}_{\bm{t}_1})) \odot \exp(-\bm{s}_1(\bm{y}_1;\boldsymbol{\theta}_{\bm{s}_1}))\, . \\
    \end{split}
\label{eq:affine_layer_backward}
\end{equation*}

To prevent numerical instabilities, we clamp and scale the exponential function as proposed in \cite{ardizzone2019guided}. Therefore, we obtain
\begin{equation}
\begin{split}
    \exp(\bm{s}_2(\bm{x}_2);\boldsymbol{\theta}_{\bm{s}_2}) &= \exp\left(\arctan\left(\bm{s}_2(\bm{x}_2);\boldsymbol{\theta}_{\bm{s}_2}\right)\right)\, , \\
    \exp(\bm{s}_1(\bm{y}_1);\boldsymbol{\theta}_{\bm{s}_1}) &= \exp\left(\arctan\left(\bm{s}_1(\bm{y}_1);\boldsymbol{\theta}_{\bm{s}_1}\right)\right)\, . \\ 
\end{split}
\label{eq:clamped_exponential}
\end{equation}

\subsection{From Invertible NNs to Invertible AEs}\label{subsec:fromINNtoINVAE}
Each invertible layer does not perform any form of dimensionality reduction because to achieve (analytical) invertibility, the layers require the same dimensionality of inputs and outputs, i.e., $\bm{x}, \bm{y} \in \mathbb{R}^N$. Therefore, using an invertible NN composed of multiple affine coupling layers is not sufficient for building an invertible AE. However, to achieve dimensionality reduction, we use a technique proposed in \cite{nguyen2019training}. We assume $\bm{x}$ to be the input of the first layer and $\bm{y}$ to be the output of the last layer of the invertible NN. In particular, after applying the forward pass transformations, we can mask with zeros the output of the final layer $\bm{y}$ up to an arbitrary latent dimension to obtain $\bm{z}$: 
\begin{equation}
\begin{split}
    \bm{y} &= \enc(\bm{x};\boldsymbol{\theta}_{\enc}) \\
    \bm{z} &= \texttt{zero-masking}(\bm{y})\, .
\end{split}
\end{equation}
In particular, given the vector $\bm{y}$, the zero-masking procedure is defined as follows:
\begin{equation}
    \begin{split}
        \bm{y} &= [y_1, \ldots,y_n, y_{n+1}, \ldots, y_N] \\
         &\downarrow \texttt{zero-masking} \\
        \bm{z} &= [y_1, \ldots, y_n, 0, \ldots, 0]\, ,
    \end{split}
\end{equation}
where $N$ indicates the dimensionality of $\bm{y}$, and $n$ the dimension of the reduced space. The zero $(N-n)$-elements of $\bm{z}$ have no attached gradient to disable backpropagation thought the corresponding weights and biases. Zero masking is an effective dimensionality-reduction mechanism, forcing the encoder network to compress data into $n$-dimensional vectors, while maintaining the output dimensionality requirement imposed by the invertible layers.
Then, the zero-masked vector $\bm{z}$ is passed through the backward pass to recover the reconstruction of the input $\hat{\bm{x}}$: 
\begin{equation}
    \hat{\bm{x}} = \dec(\bm{z};\boldsymbol{\theta}_{\dec})\, ,
\end{equation}
where $\dec =\enc^{-1}$  (see Figure \ref{fig:1a}).
The zero-masking promotes the compression of the data into a limited number of relevant dimensions, making the invertible NN and inv-AE. Analogously to traditional AEs, we train our inv-AE to minimize the reconstruction loss
\begin{equation}
\begin{split}
      \mathcal{L}(\boldsymbol{\theta}_{\enc}, \boldsymbol{\theta}_{\dec}) =  \frac{1}{M}\sum_{i=1}^M ||\bm{x}^{(i)} - \dec(\enc(\bm{x}^{(i)}; \boldsymbol{\theta}_{\enc}); \boldsymbol{\theta}_{\dec}) ||^2_2\, ,
\end{split}
\label{eq:recon_loss_invae}
\end{equation}
where encoder and decoder correspond to the forward and backward pass through the invertible layers, and that $\boldsymbol{\theta}_{\dec}=\boldsymbol{\theta}_{\enc}$.

In addition to the reconstruction loss, it is possible to utilize additional terms such a zero-masking regularization loss \cite{nguyen2019training}, or a Lipschitz-constant regularization loss \cite{behrmann2021understanding} to improve numerical stability and convergence properties of the architecture. However, in our case, the choice of an affine coupling layer with clamped exponential (see Equation \eqref{eq:clamped_exponential}) is sufficient for preventing numerical instabilities. As shown in Figure \ref{fig:spectral_norm_ablation} in \ref{sec:spectral_normalization}, spectral normalization \cite{miyato2018spectral} is only mildly beneficial for our inv-AE, demonstrating the intrinsic numerical stability of our architecture. Implementation details of the proposed  architecture are discussed in \ref{sec:nn_architecture}.

\subsection{Invertible (POD) Deep Learning-based ROMs}\label{subsec:invDLROM}

In this work, we do not only test the dimensionality-reduction capabilities of the inv-AE, but we also empirically evaluate the advantages of inv-AE in the context of projection-based ROM. Therefore, we propose to enhance the DL-ROM and the POD-DL-ROM (see Section \ref{sec:dl-rom}) with our inv-AE replacing the convolutional AE. We call these two approaches as \emph{inv-DL-ROM} and \emph{POD-inv-DL-ROM}.  Analogously to DL-ROM, inv-DL-ROM is composed of an inv-AE and a feedforward NN $\xirom$ predicting the latent vector $\tilde{\bm{z}}_k^{\boldsymbol{\mu}}$ from the timestep $k$ and the parameter vector $\boldsymbol{\mu}$:
\begin{equation*}
    \begin{split}
        \tilde{\bm{z}}_k^{\boldsymbol{\mu}} &= \xirom(k, \boldsymbol{\mu}; \boldsymbol{\theta}_{\xirom})\\
        \tilde{\bm{z}}_k^{\boldsymbol{\mu}} & = \texttt{zero-padding}(\tilde{\bm{z}}_k^{\boldsymbol{\mu}}) \\
        \tilde{\bm{x}}_k^{\boldsymbol{\mu}} &= \dec(\tilde{\bm{z}}_k^{\boldsymbol{\mu}};\boldsymbol{\theta}_{\xirom})\, ,
    \end{split}
\end{equation*}
where $\dec =\enc^{-1}$ and $\boldsymbol{\theta}_{\dec}=\boldsymbol{\theta}_{\enc}$. In this case, a zero-padding procedure is used to have the dimensionality of $\tilde{\bm{z}}_k^{\boldsymbol{\mu}}$ equal to $\tilde{\bm{x}}_k^{\boldsymbol{\mu}}$ as required by the inv-AE architecture and it is defined as follows:
\begin{equation}
    \begin{split}
        \tilde{\bm{z}}_k^{\boldsymbol{\mu}} &= [z_1, \ldots,z_n] \\
         &\downarrow \texttt{zero-padding} \\
        \tilde{\bm{z}}_k^{\boldsymbol{\mu}} &= [z_1, \ldots, z_n, 0, \ldots, 0]\,.
    \end{split}
\end{equation}  
Zero-padding involves concatenating $N-n$ zeros to the output of the the model $\boldsymbol{\xi}$ to obtain a vector of again dimension $N$.
On the other side, POD-inv-DL-ROM simply adds the POD (re)projection of the POD coefficients to the snapshots:
\begin{equation*}
    \begin{split}
        \tilde{\bm{z}}_k^{\boldsymbol{\mu}} &= \xirom(k, \boldsymbol{\mu}; \boldsymbol{\theta}_{\xirom}) \\
        \tilde{\bm{z}}_k^{\boldsymbol{\mu}} & = \texttt{zero-padding}(\tilde{\bm{z}}_k^{\boldsymbol{\mu}}) \\
        \tilde{\bm{h}}_k^{\boldsymbol{\mu}} &= \dec(\tilde{\bm{z}}_k^{\boldsymbol{\mu}};\boldsymbol{\theta}_{\xirom}) \\
        \tilde{\bm{x}}_k^{\boldsymbol{\mu}} &= U\tilde{\bm{h}}_k^{\boldsymbol{\mu}} \, , 
    \end{split}
\end{equation*}
where $U$ is the SVD matrix computed on the train set and $\tilde{\bm{h}}_k^{\boldsymbol{\mu}}$ is the reconstructed POD coefficient using $\xirom$ and $\dec$. Due to the parameter sharing of encoder and decoder, (POD) inv-DL-ROM does not require the latent-space regularization of traditional DL-ROM architectures (second term of Equation \eqref{eq:DL-ROM_training}) and can be trained with a simplified training objective 
\begin{equation}
\begin{split}
      \mathcal{L}_{\text{inv-DL-ROM}}(\boldsymbol{\theta}_{\enc}, \boldsymbol{\theta}_{\dec}, \boldsymbol{\theta}_{\xirom})&=\frac{1}{M}\sum_{i=1}^M||\bm{x}^{(i)} - \tilde{\bm{x}}^{(i)}||^2_2  \\
      &=\frac{1}{M}\sum_{i=1}^M||\bm{x}^{(i)} - \dec(\xirom(k^{(i)},\boldsymbol{\mu}^{(i)};\boldsymbol{\theta}_{\xirom});\boldsymbol{\theta}_{\dec})||^2_2\, .
\end{split}
\label{eq:inv-DL-ROM_training}
\end{equation}
While the DL-ROM objective in Equation \eqref{eq:DL-ROM_training} requires tuning of the coefficient $\alpha$ to balance the contribution of the two terms of the loss function, the parameters of inv-DL-ROM are optimized solely to reconstruct the data $\bm{x}$ from $\boldsymbol{\mu}$ and $t$.

\section{Numerical Results}\label{sec:results}

To investigate the capabilities on the invertible AEs, we introduce three numerical examples, namely a parametric 1-dimensional Burgers' equation (Section \ref{Sec:num_exp_burgers}), a parametric 2-dimensional fluid flow around an obstacle with variable geometry (Section \ref{Sec:num_exp_flow}), and a parametric 3-dimensional Korteweg-de Vries equation (Section \ref{Sec:num_exp_kdv}). Using these three numerical examples, we aim to test \emph{(i)} the dimensionality reduction capabilities of (POD)inv-AE 
and \emph{(ii)} the accuracy of the resulting ROM, namely (POD)inv-DL-ROM. For reconstruction, we compare different architectures:
\begin{itemize}
    \item[-] \emph{POD} -- proper orthogonal decomposition,
    \item[-] \emph{conv. AE} -- convolutional AE with $\sim 0.25M$ parameters,
    \item[-] \emph{fully-conn. AE} -- fully-connected AE and with $\sim 2.5M$, parameters.
    \item[-] \emph{inv-AE} -- invertible AE with $\sim2.5M$ parameters, and
    \item[-] \emph{inv-AE (light)} -- invertible AE with $\sim1.25M$ parameters.
\end{itemize}
For reduction, instead, we compare:
\begin{itemize}
    \item[-] \emph{POD-NN} -- proper orthogonal decomposition with neural network-based surrogate model \cite{hesthaven2018non},
    \item[-] \emph{DL-ROM (conv. AE)} -- DL-ROM with convolutional AE,
    \item[-] \emph{DL-ROM (fully-conn. AE)} -- DL-ROM with fully-connected AE,
    \item[-] \emph{inv-DL-ROM} -- DL-ROM with inv-AE, and
    \item[-] \emph{inv-DL-ROM (light)} -- DL-ROM with inv-AE (light).
\end{itemize}
For fairness of comparison, all the AEs share the same number of layers, and the same architecture for $\xirom(k, \boldsymbol{\mu}; \boldsymbol{\theta}_{\xirom})$, composed of 2 hidden layers with 256 neurons each and leaky ReLU activations. For the 2-dimensional and 3-dimensional test cases, a POD step with $N \gg r\gg n$ is included before the AEs to reduced the data dimensionality without compromising the accuracy of the reconstructions and consequently reducing the number of learnable parameters of the neural network-based models. The complete architectural and implementation details are discussed in \ref{sec:nn_architecture}.

After training the models using the train set $\bm{X}_{\text{train}}$, we evaluate their performance on the test set $\bm{X}_{\text{test}}$, where we set $N_{\textnormal{test}}:= |\mathcal{P}_{\text{test}} |$.
\begin{equation*}
  \bm{X}_{\text{test}} =\left[ \begin{array}{ccccc} | & | & \cdots & | & | \\
         {\bm{x}_{1}}^{\boldsymbol{\mu}_1} & {\bm{x}}_{2}^{\boldsymbol{\mu}_1} & \cdots & {\bm{x}}_{N_t - 1}^{\boldsymbol{\mu}_{N_{\textnormal{test}}}}  &  {\bm{x}}_{N_t}^{\boldsymbol{\mu}_{N_{\textnormal{test}}}} \\
         | & | & \cdots & | & |   \end{array} \right] \in \mathbb{R}^{N \times (N_t \times N_{\text{test}})} \, .
\label{eq:test_set}
\end{equation*}
In particular, we utilize the mean reconstrution error given by
\begin{equation}
\frac{1}{|\bm{X}_{\text{test}}|} \sum_{\boldsymbol{\mu}\in \mathcal{P}_{\text{test}}} \sqrt{\frac{\sum_{k=1}^{N_t}||\bm{x}^{\boldsymbol{\mu}}_k  - \hat{\bm{x}}^{\boldsymbol{\mu}}_k||_2^2}{\sum_{k=1}^{N_t}||\bm{x}^{\boldsymbol{\mu}}_k||_2^2}}\, ,
\label{eq:proj_error}
\end{equation}
and the mean reduction error 
\begin{equation} 
\frac{1}{|\bm{X}_{\text{test}}|} \sum_{ \boldsymbol{\mu}\in \mathcal{P}_{\text{test}}} \sqrt{\frac{\sum_{k=1}^{N_t}||\bm{x}^{\boldsymbol{\mu}}_k - \tilde{\bm{x}}^{\boldsymbol{\mu}}_k  ||_2^2}{\sum_{k=1}^{N_t}||\bm{x}^{\boldsymbol{\mu}}_k||_2^2}}\, ,
\label{eq:red_error}
\end{equation} 
where $\bm{x}^{\boldsymbol{\mu}}_k$ indicate the discretized FOM snapshot at the timestep $k$, $\hat{\bm{x}}^{\boldsymbol{\mu}}_k$ the reconstructed snapshot when passed through encoder and decoder, and $\tilde{\bm{x}}^{\boldsymbol{\mu}}_k$ the decoded solution of the DL-ROM. The reconstruction error measures the best-possible error made by the corresponding chosen autoencoder realization, while the reduction error shows the error of the corresponding ROM.

\subsection{Burgers' Equation}\label{Sec:num_exp_burgers}

We study a time-dependent 1-dimensional Burgers' equation as introduced in \cite{lee2020model} given by
\begin{equation*}
\begin{split}
        \frac{\partial x(y,t; \boldsymbol{\mu})}{\partial t} + \frac{\partial f(x(y,t; \boldsymbol{\mu}))}{\partial y} - 0.02 \exp^{\mu_{2}y} &= 0, \ \ \ \forall y \in [0, 100], \forall t \in (t_0, t_f], \\
        x(0, t; \boldsymbol{\mu}) &= \mu_1, \ \ \ \forall t \in [t_0, t_f], \\
        x(y, 0) &= 1, \ \ \ \forall y \in [0, 100], \\  
\end{split}
\end{equation*}
where the flux is $f(x)=0.5x^2$ and $\boldsymbol{\mu}=(\mu_1, \mu_2)$ is the parameter vector, $t$ indicates the time variable, and $y$ the spatial variable. Analogously to \cite{lee2020model}, we derive the FOM snapshpts by numerically solving the Burgers' equation from $t_0=0.0$, to $t_f=35$, where we use equidistant time steps of $\Delta t=0.07$s resulting in $N_t=500$ and discretize in time with the implicit Euler scheme. For the spatial discretization, we apply a Godunov's scheme with $N=256$ spatial degrees of freedom. The parameter domain is $\mathcal{P}= [4.25, 5.5] \times [0.015, 0.03]$, where we choose a training set of parameters $\mathcal{P}_{\text{train}}$ by discretizing the parameter intervals into $10\times 8 = 80$ bins to replicate \cite{lee2020model}. 
The training set $X_{\text{train}}$ is composed of the 80 FOM solutions generated for the parameters $\mathcal{P}_{\text{train}}$. Further, we keep 8 of the 80 FOM solutions for the validation set $X_{\text{valid}}$, while the test set $X_{\text{test}}$ is composed of 8 FOM solutions generated with 8 unseen parameter values. In this way, we are able to test the generalization capabilities of the proposed AE architectures on unseen instances of the parameter $\boldsymbol{\mu}$. 

In this first experiment, we compare the reconstruction error achieved by our inv-AE and the convolutional AE introduced in \cite{lee2020model} for different manifold dimensions $ \in \{1, 2, 3, 4, 5, 10, 20, 50, 100, 150, 200\}$. In the second experiments, we compare instead the reduction error achieved by inv-DL-ROM with respect to DL-ROM (with the convolution AE of \cite{lee2020model} and with a fully connected AE with similar amounts of parameters) for different dimensions of the latent space  $\in \{1, 2, 3, 4, 5, 10, 20\}$. For fairness of the comparison, we train the different models for 1000 epochs with the AdamW optimizer \cite{loshchilov2017decoupled}. In addition, we use the same learning rate of $1e^{-4}$ and weight decay of $1e^{-4}$, and we employ early stopping if the loss on the validation set does not decrease after $500$ epochs of training. Same settings were used in \cite{lee2020model}. After training the models using two independent seeds, we analyze the reconstruction errors achieved by the different AE architectures and the reduction errors achieved by the different DL-ROM architectures.

\begin{figure}[h!]
    \centering
    \includegraphics[width=0.7\linewidth]{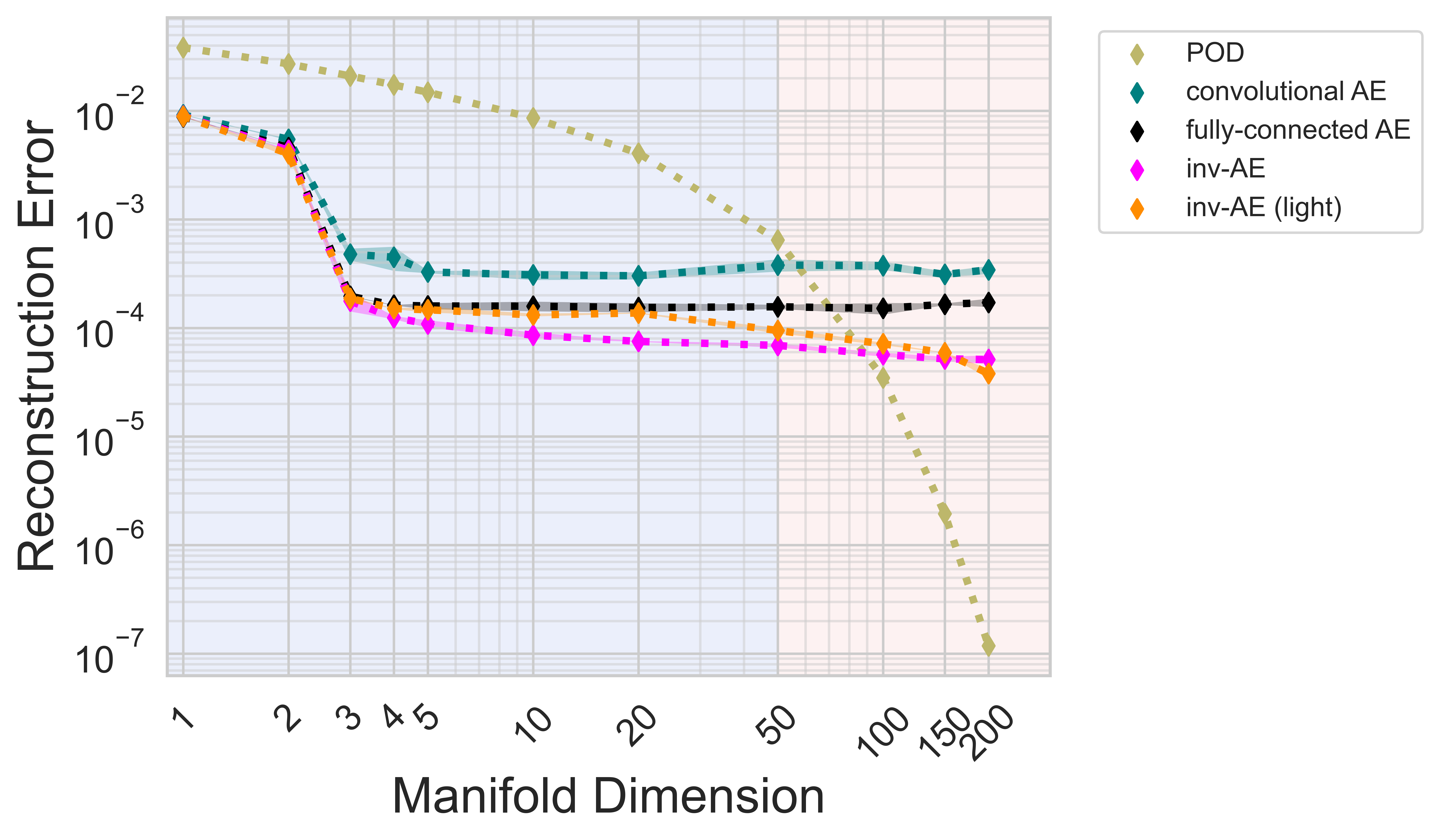}
    \caption{Reconstruction errors \eqref{eq:proj_error} calculated on the test set $
    \bm{X}_{\text{test}}$ for different dimensions of the trial manifold. The dashed line represents the mean and the shaded area the standard deviation across two different random seeds.}
    \label{fig:burgers_projection_error}
\end{figure}

In Figure \ref{fig:burgers_projection_error} we show the reconstruction errors obtained with the different architectures with respect to the chosen dimension of the trial manifold and in Table \ref{tab:projection_errors_burgers} we report the best values. We compare the reconstruction error of the inv-AEs with POD, the convolutional AE from \cite{lee2020model}, and a fully connected AE. The blue-shaded area highlights values up to manifold dimension equal to $50$, while the red-shaded area showed the region where the POD achieves comparable or superior performance to the AEs.
As reported in \cite{lee2020model}, the convolutional AE exhibits a plateau of the reconstruction error when the dimension of the manifold is $5$. A similar plateau  is achieved by the fully connected AE, while the reconstruction error of our inv-AEs continuously decreases with the increment of the manifold dimension and it is consistently smaller than the one of the convolutional AE for a manifold dimension $>2$. The POD has higher reconstruction error than both variants of the AEs for manifold dimensions smaller that $50$, while above $100$ it outperforms all AEs.

\begin{figure}[h!]
    \centering
    \includegraphics[width=0.7\linewidth]{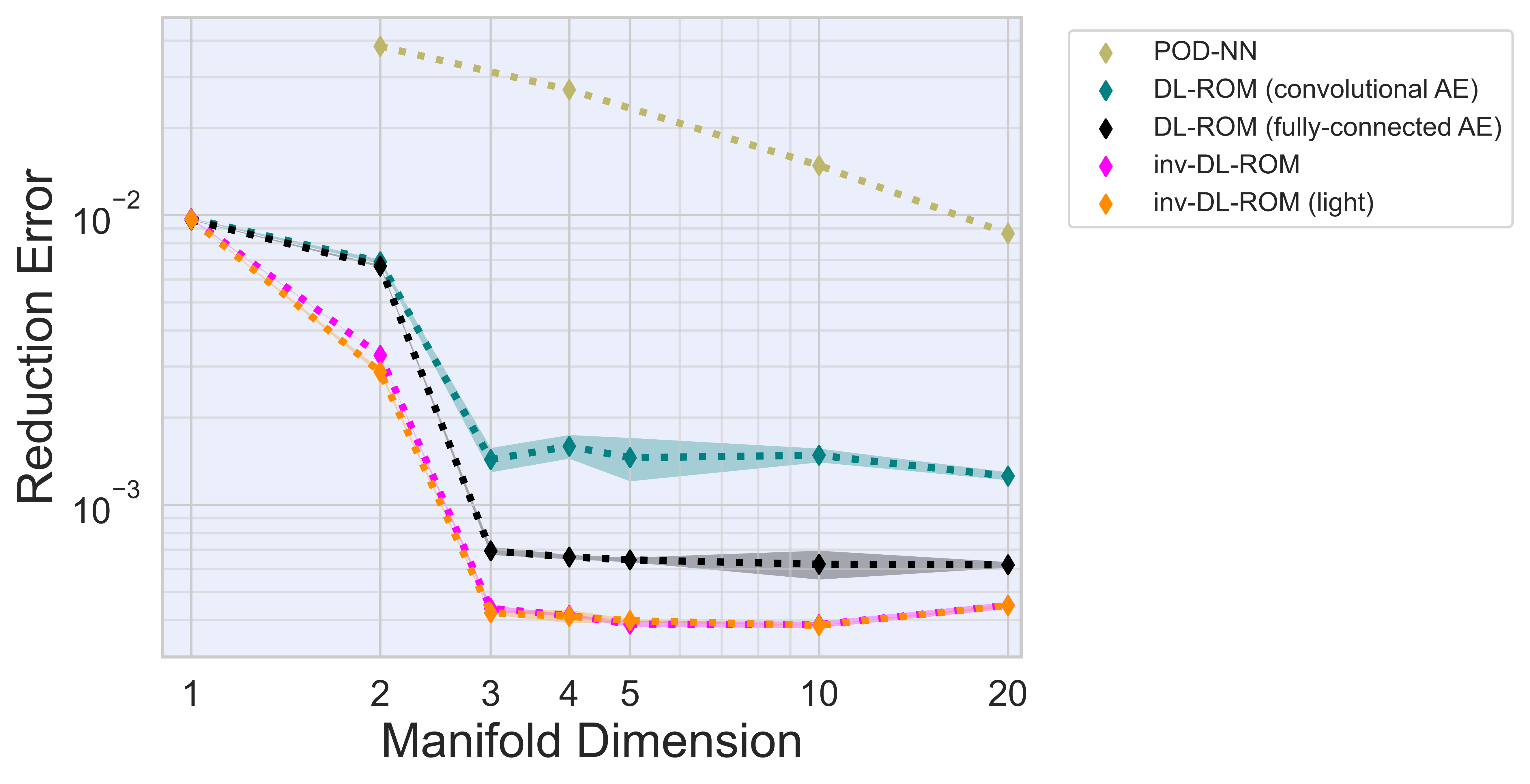}
    \caption{Reduction errors \eqref{eq:red_error}  calculated on the test set $
    \bm{X}_{\text{test}}$ for different dimensions of the trial manifold of DL-ROMs and inv-DL-ROMs. The dashed line represents the mean and the shaded area the standard deviation across two different random seeds.} 
    \label{fig:burgers_reduction_error}
\end{figure}

Further, we report in Figure \ref{fig:burgers_reduction_error} the reduction errors achieved by DL-ROMs and inv-DL-ROMs for different dimension of the trial manifold $\in \{ 1, 2,3, 4, 5, 10, 20\}$ and in Table \ref{tab:reduction_errors_burgers} we report the best values. The dimensions of the manifold were chosen to balance reduction capabilities and accuracy of the learned ROMs. By solely replacing the convolutional AE with the inv-AE, the inv-DL-ROM is capable of improving the reduction errors approximately by a factor $2.5$. We also show that all the AEs achieves superior performance to POD-NN.

\begin{figure*}[h!]
    \centering
    
    \begin{sideways} \makebox[0pt][l]{\hspace{-0.7cm} \shortstack[c]{{\bf \footnotesize FOM solution}}} \end{sideways}
    \begin{minipage}{0.95\linewidth}
    \centering
    \hspace{-0.12cm}
    \subfloat{\includegraphics[width=0.45\linewidth]{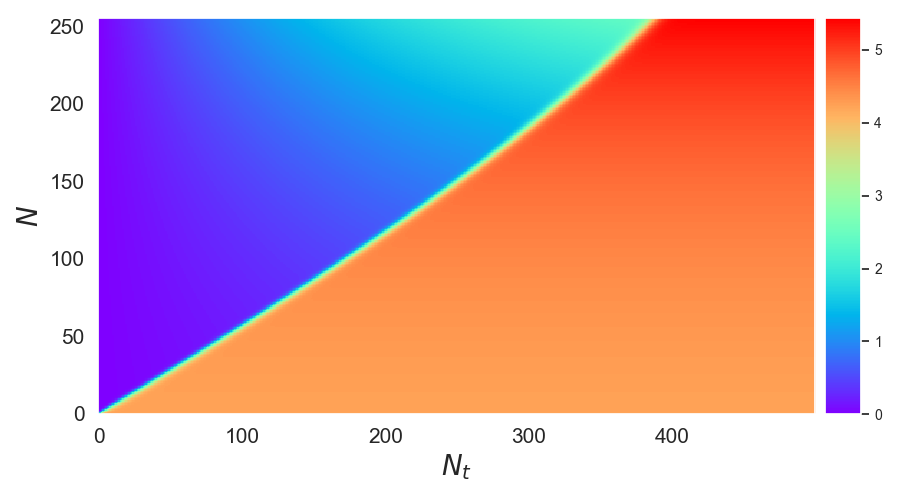}}
    \subfloat{\includegraphics[width=0.45\linewidth]{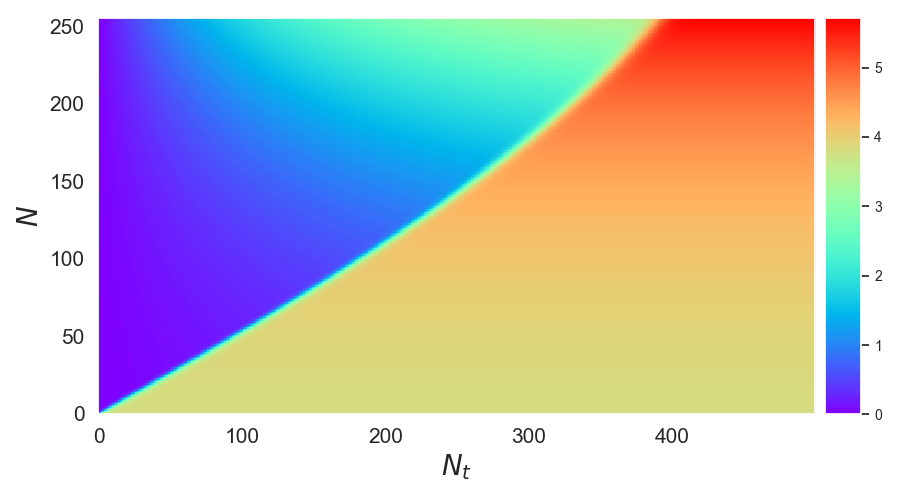}}
    \end{minipage}
    \hspace{-0.35cm}
    
    \begin{sideways} \makebox[0pt][l]{\hspace{-0.7cm} \shortstack[c]{{\bf \footnotesize Conv. AE}}} \end{sideways}
    \begin{minipage}{0.95\linewidth}
    \centering
    \hspace{-0.12cm}
    \subfloat{\includegraphics[width=0.45\linewidth]{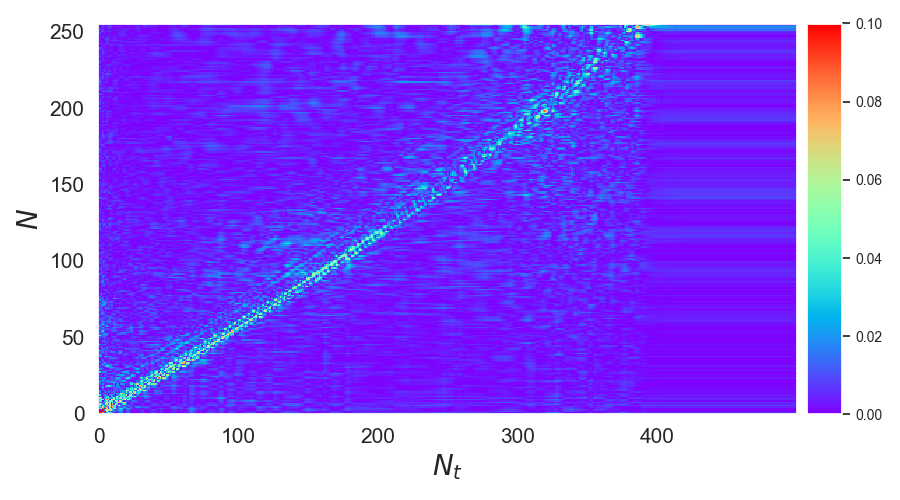}}
    \subfloat{\includegraphics[width=0.45\linewidth]{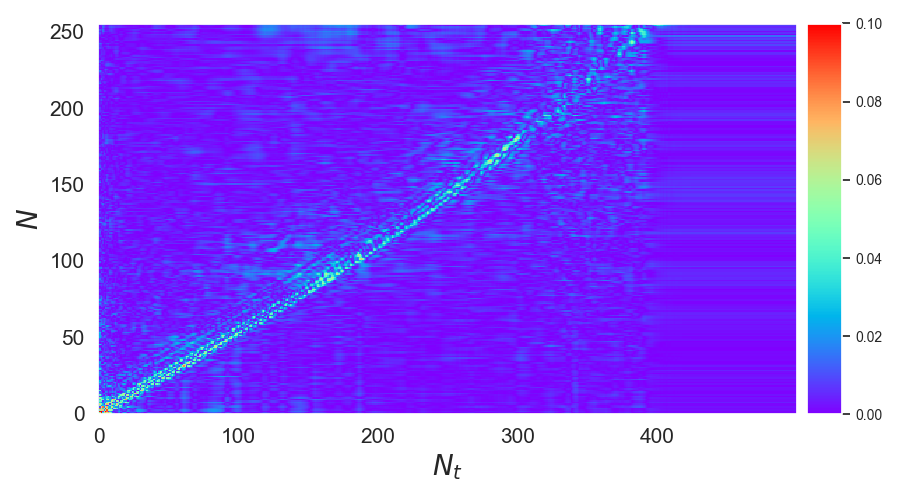}}
    \end{minipage}
    \hspace{-0.35cm}

    \begin{sideways} \makebox[0pt][l]{\hspace{-1.0cm} \shortstack[c]{{\bf \footnotesize Fully-Conn. AE}}} \end{sideways}
    \begin{minipage}{0.95\linewidth}
    \centering
    \hspace{-0.12cm}
    \subfloat{\includegraphics[width=0.45\linewidth]{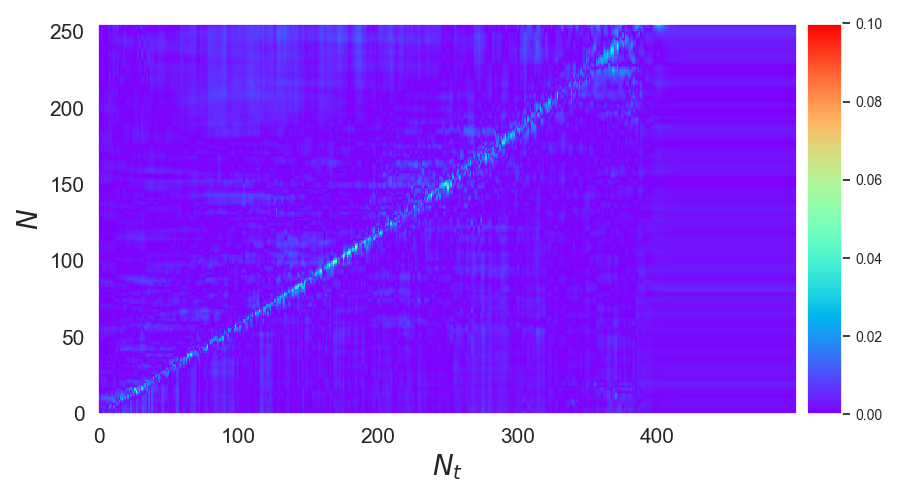}}
    \subfloat{\includegraphics[width=0.45\linewidth]{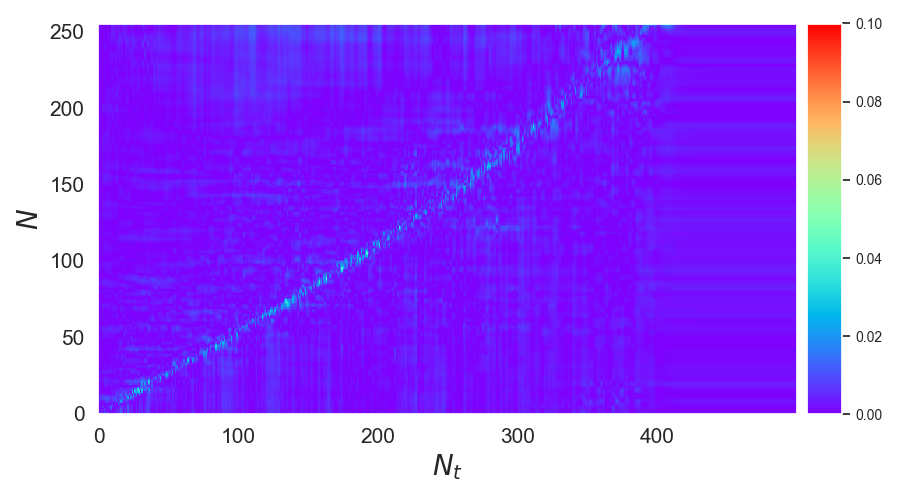}}
    \end{minipage}
    \hspace{-0.35cm}

    \begin{sideways} \makebox[0pt][l]{\hspace{-0.4cm} \shortstack[c]{{\bf \footnotesize Inv-AE}}} \end{sideways}
    \begin{minipage}{0.95\linewidth}
    \centering
    \hspace{-0.12cm}
    \subfloat{\includegraphics[width=0.45\linewidth]{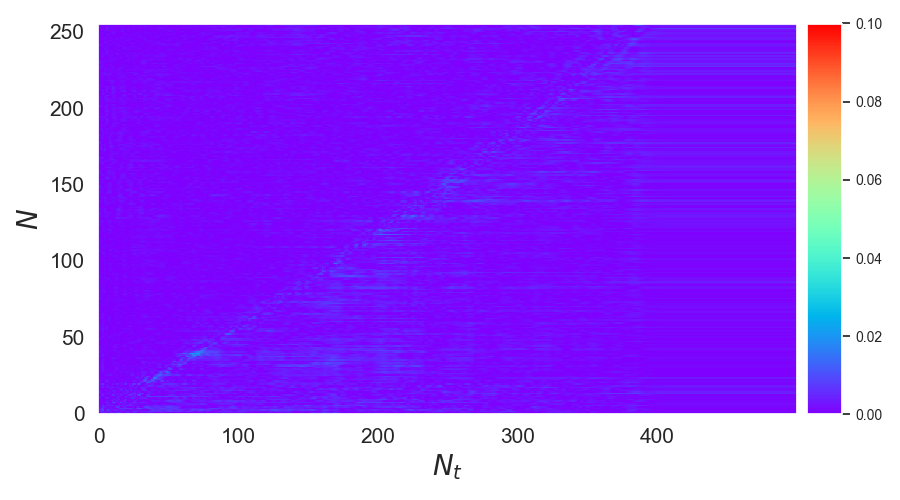}}
    \subfloat{\includegraphics[width=0.45\linewidth]{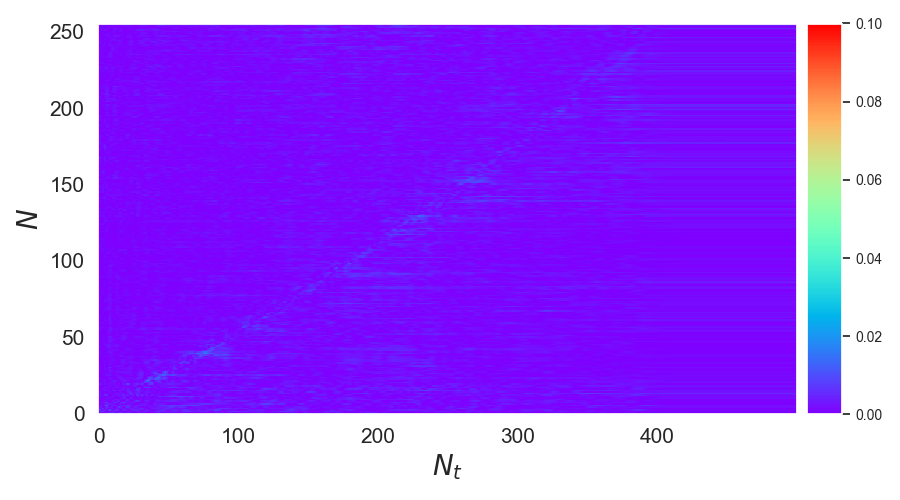}}
    \end{minipage}
    \hspace{-0.35cm}

    \begin{sideways} \makebox[0pt][l]{\hspace{-0.7cm} \shortstack[c]{{\bf \footnotesize Inv-AE (light)}}} \end{sideways}
    \begin{minipage}{0.95\linewidth}
    \centering
    \hspace{-0.12cm}
    \subfloat{\includegraphics[width=0.45\linewidth]{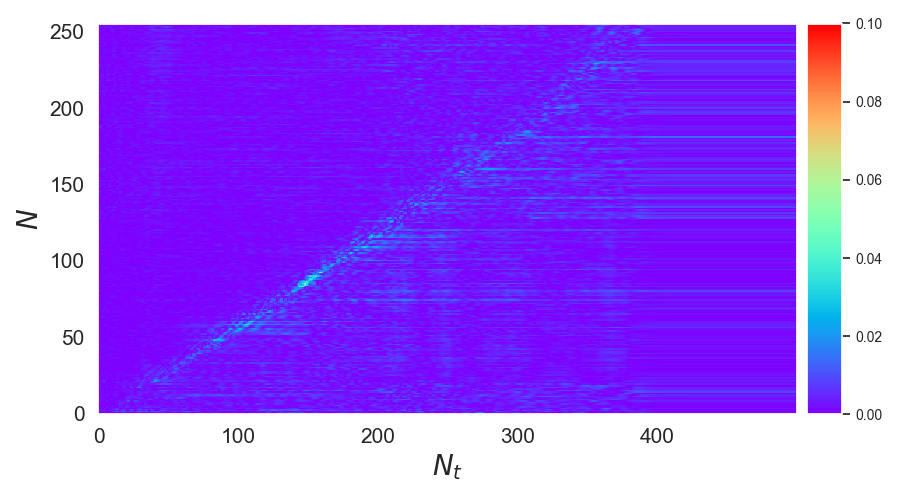}}
    \subfloat{\includegraphics[width=0.45\linewidth]{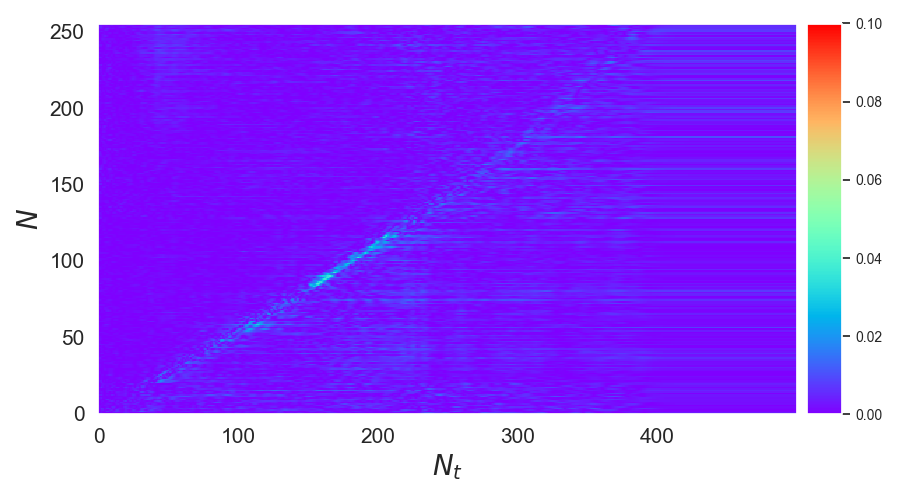}}
    \end{minipage}
    \hspace{-0.35cm}
    
    \vspace{-0.3cm}
    \caption{Qualitative comparison of the different architecture for a fixed manifold dimension of 20. The first row shows two FOM solutions from the test set, while the remaining rows the absolute reconstruction error achieved by the different AEs.}
    \label{fig:burgers_solutions}
\end{figure*}

Eventually, in Figure \ref{fig:burgers_solutions} we show a representative example of reconstruction errors of the different AE architectures for a manifold dimension of 20, where it is possible to notice the improved reconstruction capabilities of the inv-AEs compared to convolutional and fully-connected AEs.

\subsection{Flow Around an Obstacle} \label{Sec:num_exp_flow} 

\begin{figure}
    \centering
    \includegraphics[width=1.0\linewidth]{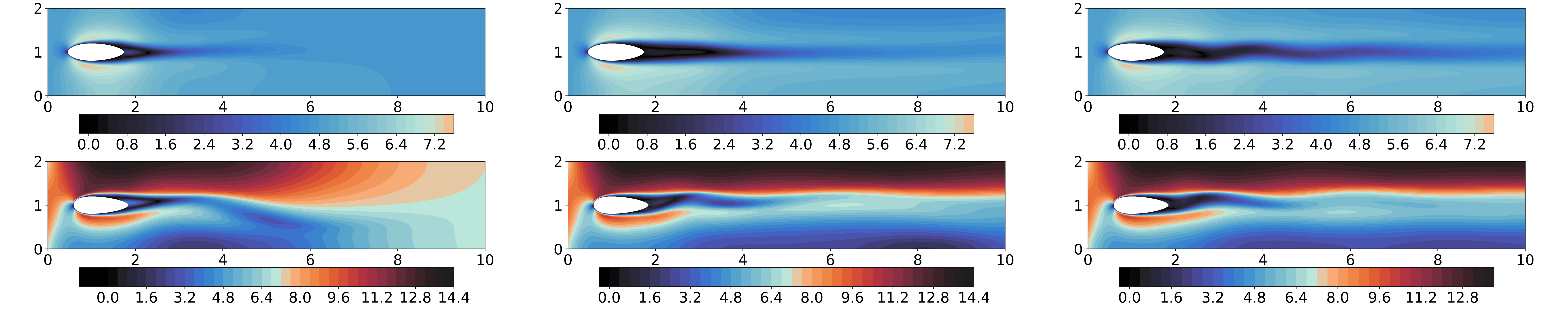}
    \caption{Examples of magnitude of the velocity fields of the flow around an obstacle at different timesteps and parameter instances.}
    \label{fig:examples_solutions_flow}
\end{figure}

As a second challenging and high-dimensional test case, we use the fluid flow around an obstacle. We consider a 2-dimensional channel of dimension $[0,10]\times[0,2]$ with a circular obstacle placed in $(1,1)$ with a radius of $0.2$. The obstacle can be deformed along the x-axis on its left and right side by $\gamma_l \in [0.2, 0.6]$ and $\gamma_r \in [0.2, 1.0]$, respectively. We also assume incompressible flow, whose dynamics is described by the unsteady Navier-Stokes equations
\begin{equation}
\begin{cases}
       \dfrac{\partial \bm{v}(\bm{y},t)}{\partial t} -\mu \Delta \bm{v}(\bm{y},t) + (\bm{v}(\bm{y},t) \cdot \nabla) \bm{v}(\bm{y},t) + \nabla p(\bm{y},t) = 0  \qquad &\mathrm{in} \ \Omega(\boldsymbol{\mu}) \times (t_0,t_f) \\
       \mathrm{div }\  \bm{v}(\bm{y},t) = 0  \qquad &\mathrm{in} \ \Omega(\boldsymbol{\mu}) \times (t_0,t_f) \\
       \bm{v}(\bm{y},t) = \bm{v}_{\text{in}}(\bm{y},t;\boldsymbol{\mu})  \qquad &\mathrm{on} \ \Gamma_{\textrm{in}} \times (t_0,t_f)\\
       \bm{v}(\bm{y},t) = \boldsymbol{0} \qquad &\mathrm{on} \ \Gamma_{\textrm{obs}}(\boldsymbol{\mu}) \times (t_0,t_f)\\
       \bm{v}(\bm{y},t) \cdot \bm{n}(\bm{y}) = 0  \qquad &\mathrm{on} \ \Gamma_{\textrm{walls}} \times (t_0,t_f) \\
       (\mu \nabla \bm{v}(\bm{y},t) - p(\bm{x},t)I)\bm{n}(\bm{y}) \cdot \boldsymbol{\tau}(\bm{x}) = 0  \qquad &\mathrm{on} \ \Gamma_{\textrm{walls}} \times (t_0,t_f) \\
       (\mu \nabla \bm{v}(\bm{y},t) - p(\bm{y},t)I)\bm{n}(\bm{x}) = 0  \qquad &\mathrm{on} \ \Gamma_{\textrm{out}} \times (t_0,t_f)\\
       \bm{v}(\bm{y},0) = \boldsymbol{0} &\mathrm{in} \ \Omega(\boldsymbol{\mu}) \times \{t = t_0\}
\end{cases}
    \label{eq:ns}
\end{equation}  
where $\Omega(\boldsymbol{\mu})=\Omega(\gamma_l, \gamma_r)$ is the domain of interest depending on the parameters $\gamma_l$ and $\gamma_r$, $t_0=0$ is the initial time, $t_f > 0$ is the final time, $\bm{y}=(y_1, y_2)$ indicates the spatial variables, $\bm{v}(\bm{y},t): \Omega(\gamma_l, \gamma_r) \times [0,t_f] \to \mathbb{R}^2$ is the velocity, $p(\bm{y},t): \Omega(\gamma_l, \gamma_r) \times [0,t_f] \to \mathbb{R}$ is the pressure, $\bm{v}_{\text{in}}(\bm{y},t;\boldsymbol{\mu}): \Gamma_{\textrm{in}} \times [0,t_f] \to \mathbb{R}^2$ is the time and parameters-dependent inflow datum, $\mu=0.01$ is the kinematic viscosity, $\bm{n}(\bm{y})$ and $\boldsymbol{\tau}(\bm{y})$ are the normal and tangential tensors.
The fluid starts with zero velocity at $t_0=0$ and enters the domain from the left boundary with intensity $\gamma_{\text{in}} \in [1.0, 10.0]$ and angle of attack $\alpha_{\text{in}} \in [-1.0, 1.0]$ as defined in
\begin{equation*}
    \bm{v}((0, y_2), t) = (\gamma_{\text{in}}\cos{\alpha_{\text{in}}}, y_2(2-y_2)\gamma_{\text{in}}\sin{\alpha_{\text{in}}})\, ,
\end{equation*}
where the parabolic profile $y_2(2-y_2)$ is used to prevent discontinuities as in \cite{tomasetto2025reducedordermodelingshallow}. We use on a mesh with $N=80592$ degrees of freedom for the velocities and solve \eqref{eq:ns} through the
incremental Chorin-Temam reconstruction method for $t_f=5$s, and $\Delta t=0.05$s using the finite element solver in \texttt{fenics} \cite{alnaes2015fenics} (see Figure \ref{fig:examples_solutions_flow}). 

Our goal is to recover the velocity field $\bm{v}$ for different values of the parameters $\boldsymbol{\mu}(\gamma_l, \gamma_r, \gamma_{\text{in}}, \alpha_{\text{in}})$. The parameter domain is $\mathcal{P}= [0.2, 0.6] \times [0.2, 1.0] \times [1.0, 10.0] \times [-1.0, 1.0]$, corresponding to Reynolds numbers in the range $[40, 1600]$, where we choose a training set of parameters $\mathcal{P}_{\text{train}}$ by randomly sampling 80 times from $\mathcal{P}$, the validation set of parameters $\mathcal{P}_{\text{valid}}$  by randomly sampling 10 times from $\mathcal{P}$, and the test set of parameters $\mathcal{P}_{\text{test}}$  by randomly sampling 10 times from $\mathcal{P}$. 
Therefore, the training set $X_{\text{train}}$ is composed of the 80 FOM solutions generated for the parameters $\mathcal{P}_{\text{train}}$. Further, we keep 10 FOM solutions for the validation set $X_{\text{valid}}$, while the test set $X_{\text{test}}$ is composed of 10 FOM solutions generated with 10 unseen parameter values.

\begin{figure}[h!]
    \centering
    \includegraphics[width=0.7\linewidth]{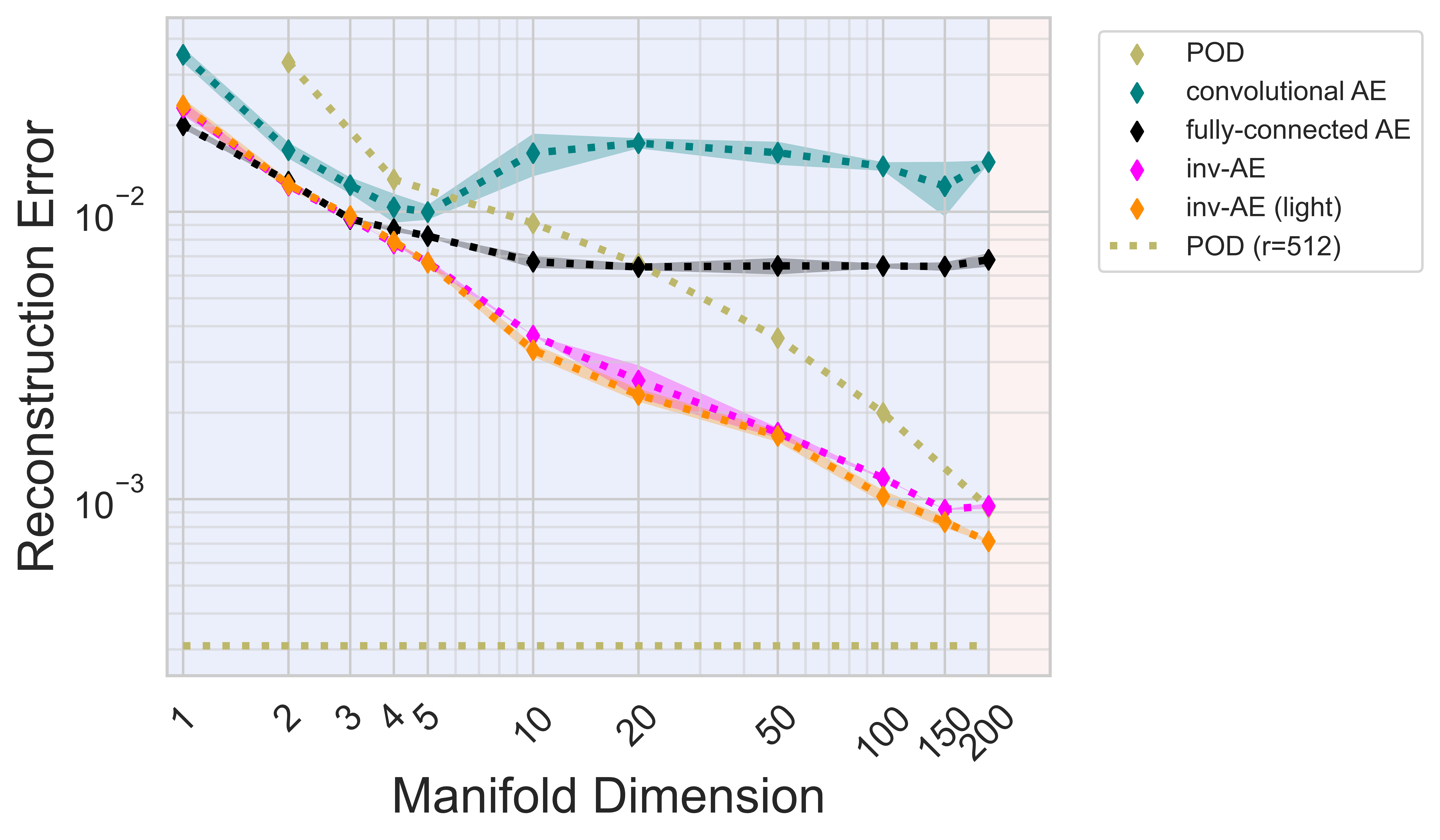}
    \caption{Reconstruction errors \eqref{eq:proj_error} for different dimensions of the trial manifold. The dashed line represents the mean and the shaded area the standard deviation across two different random seeds.}
    \label{fig:flow_projection_error}
\end{figure}

Due to the high dimensionality of the fluid velocities, we first apply the POD to the snapshots, treated as a vector, to reduce the dimensionality from $80592$ to $r=512$, namely $256$ dimensions per velocity component, posing a lower bound on the reconstruction error of $3.0 e^{-4}$. It is worth highlighting that with the POD pre-processing step we do not aim to find the smallest number of basis functions, but simply to decrease the snapshot dimensionality to the point in which we can use AEs with limited numbers of learnable parameters that are less computationally demanding to train and  do not require larger amounts of data. A similar procedure was used in \cite{fresca2022pod}. From Figure \ref{fig:flow_projection_error}, it is possible to notice that no AEs is capable of getting close to the lower bound posed by the POD. Therefore, the choice of considering $r=512$ dimension is not detrimental to the AE and can be considered suitable for this numerical example. We compare our inv-AEs with the convolutional AE introduced in \cite{fresca2021comprehensive} for a similar test case and a fully connected AE. 
We train the different AE models for 1500 epochs with AdamW optimizer with different manifold dimensions $\in \{1, 2, 3, 4, 5, 10, 20, 50, 100\}$. For all the models, we use learning rate equal to  $1e^{-4}$, weight decay equal to  $1e^{-4}$, and we employ early stopping if the validation loss does not decrease after $500$ epochs of training.

In Figure \ref{fig:flow_projection_error}, we show the reconstruction error \eqref{eq:proj_error} obtained using the convolutional AE, fully connected AE, inv-AEs, and POD, and in Table \ref{tab:projection_errors_flow} we report the best values. In addition, we report the reconstruction error when truncating the POD to $r=512$. Even in this case, both inv-AE versions reduce the reconstruction error compared to the convolutional AE and fully connected AE. Moreover, for both of the inv-AE, we observe an almost monotonic decay of the reconstruction error, which is not the case for the convolutional or fully connected AE.  

\begin{figure}
    \centering
    \includegraphics[width=0.7\linewidth]{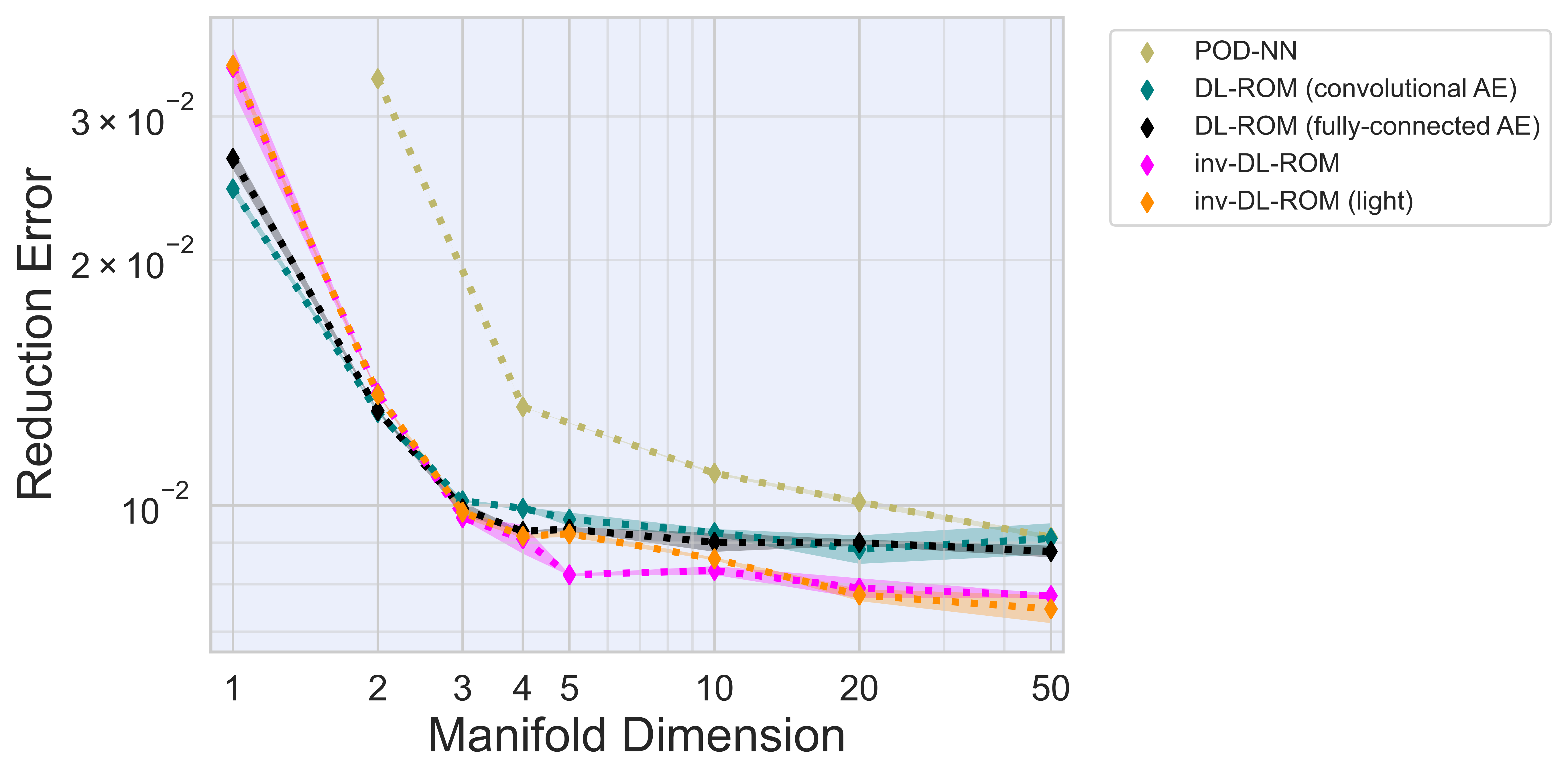}
    \caption{Reduction errors \eqref{eq:red_error} of DL-ROMs and inv-DL-ROMs. The dashed line represents the mean and the shaded area the standard deviation across two different random seeds.}
    \label{fig:flow_reduction_error}
\end{figure}

Finally, in Figure \ref{fig:flow_reduction_error}, we show the reduction errors \eqref{eq:red_error} of POD-DL-ROMs and the POD-inv-DL-ROMs and in Table \ref{tab:reduction_errors_flow} we report the best values. Even in this case, by simply replacing the convolutional or fully connected AE with the inv-AEs, we are capable of mildly improving the reduction error of the POD-DL-ROM architecture. 

\subsection{Korteweg-de Vries} \label{Sec:num_exp_kdv} 

\begin{figure}[t]
\centering
    \begin{subfigure}[b]{0.32\textwidth}
        \centering
        \includegraphics[width=\textwidth]{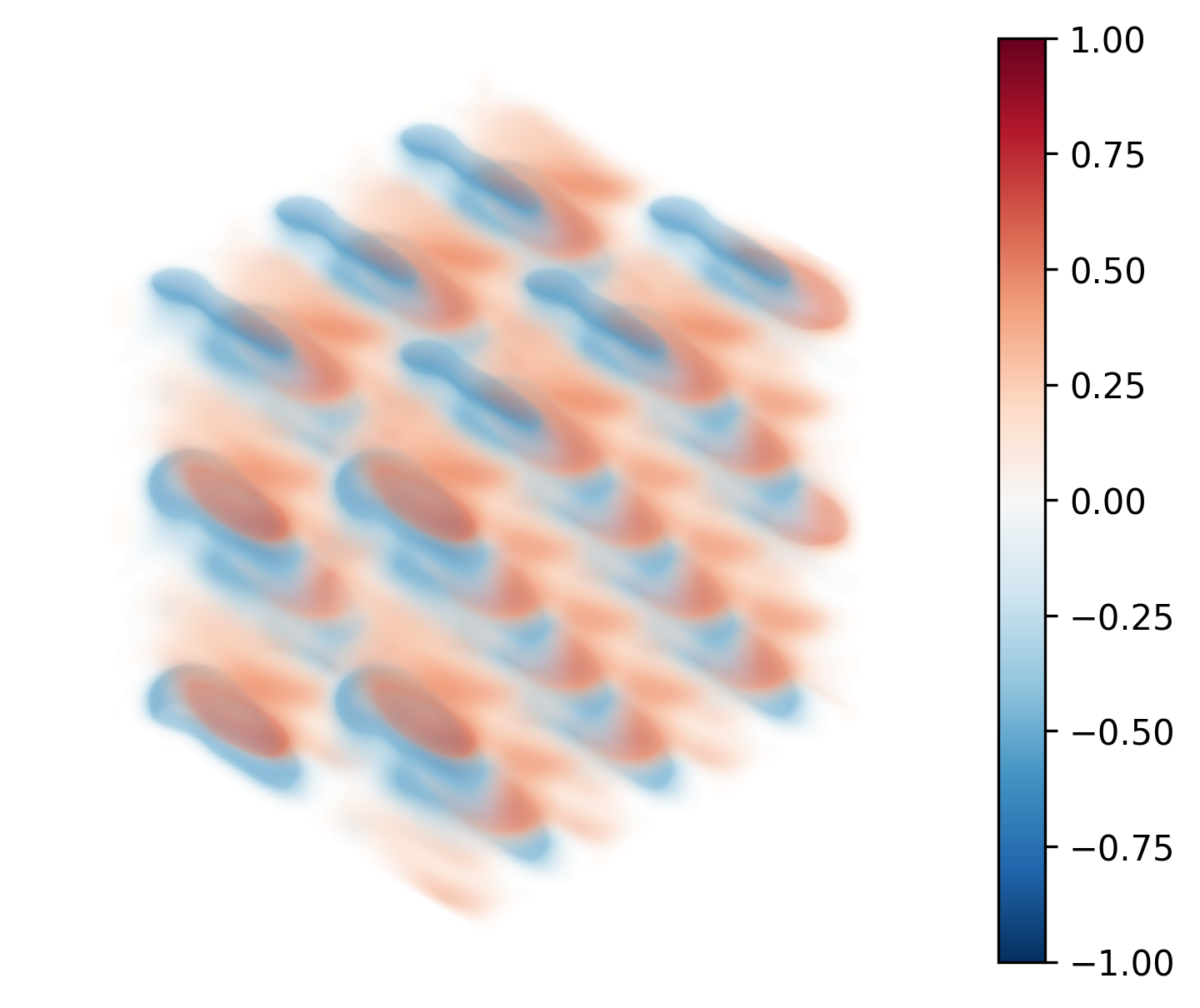}
        \label{}
     \end{subfigure}
    \begin{subfigure}[b]{0.32\textwidth}
         \centering
          \includegraphics[width=\textwidth]{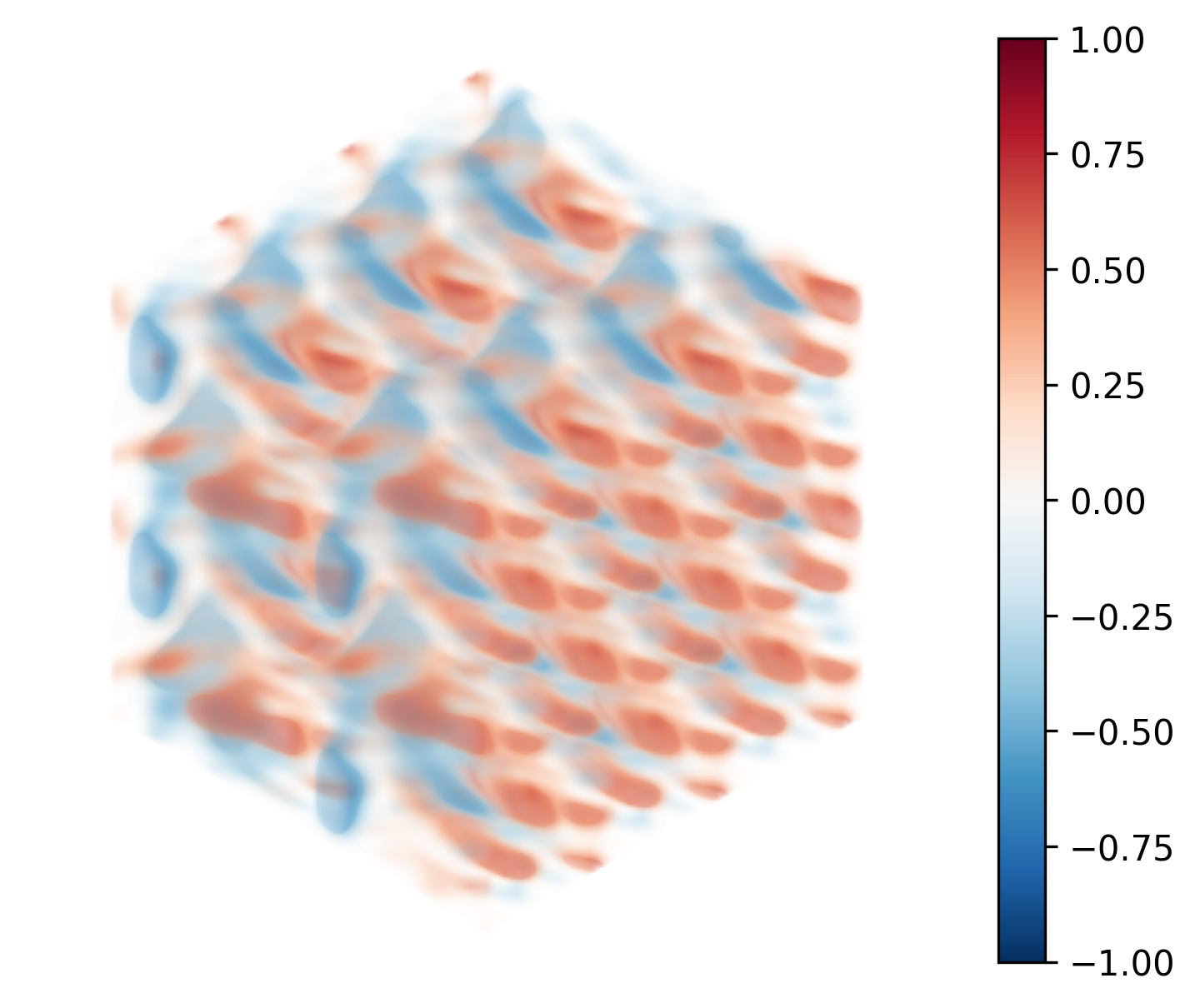}
         \label{}
     \end{subfigure}
         \begin{subfigure}[b]{0.32\textwidth}
         \centering
          \includegraphics[width=\textwidth]{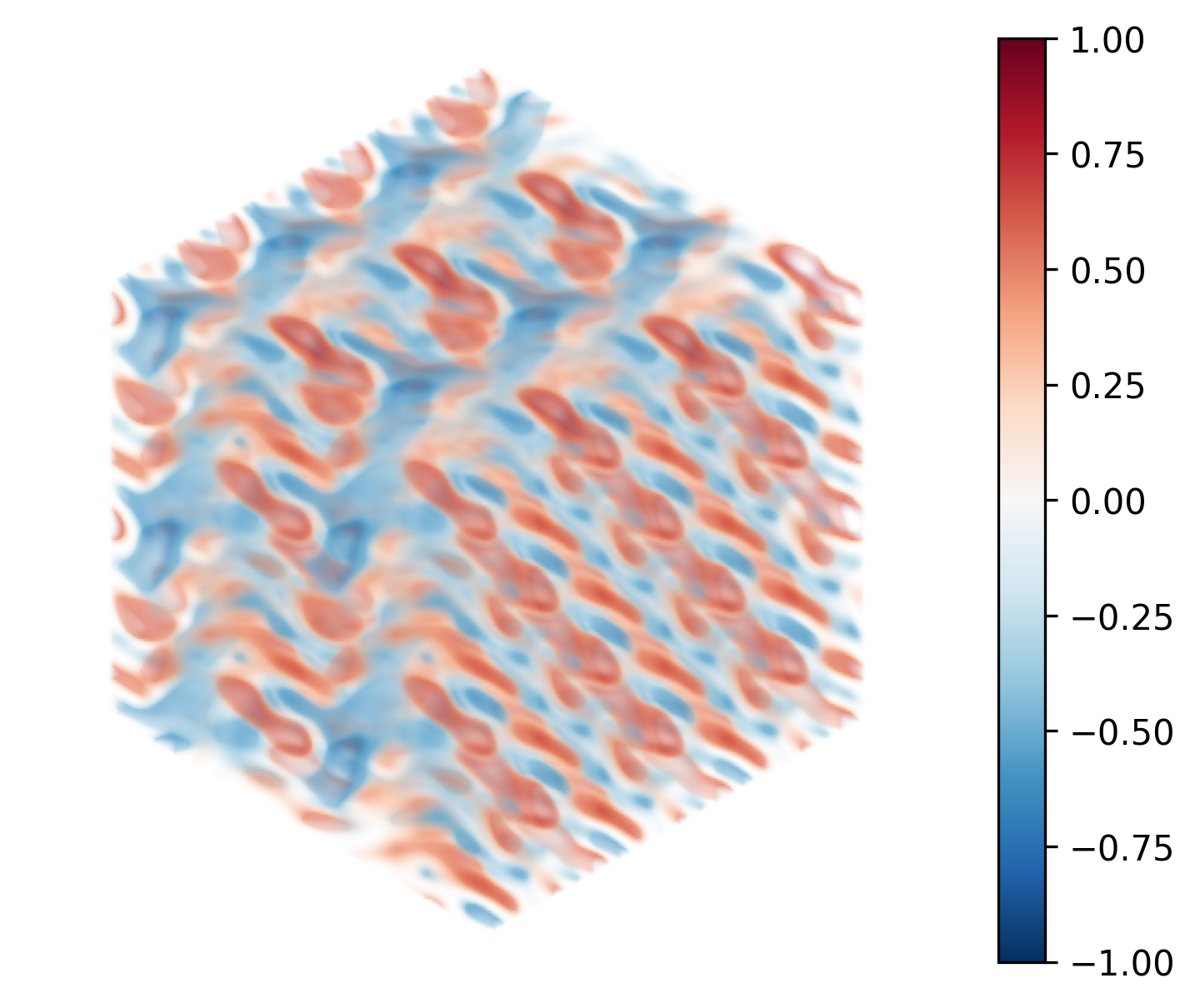}
         \label{}
     \end{subfigure}
         \begin{subfigure}[b]{0.32\textwidth}
        \centering
        \includegraphics[width=\textwidth]{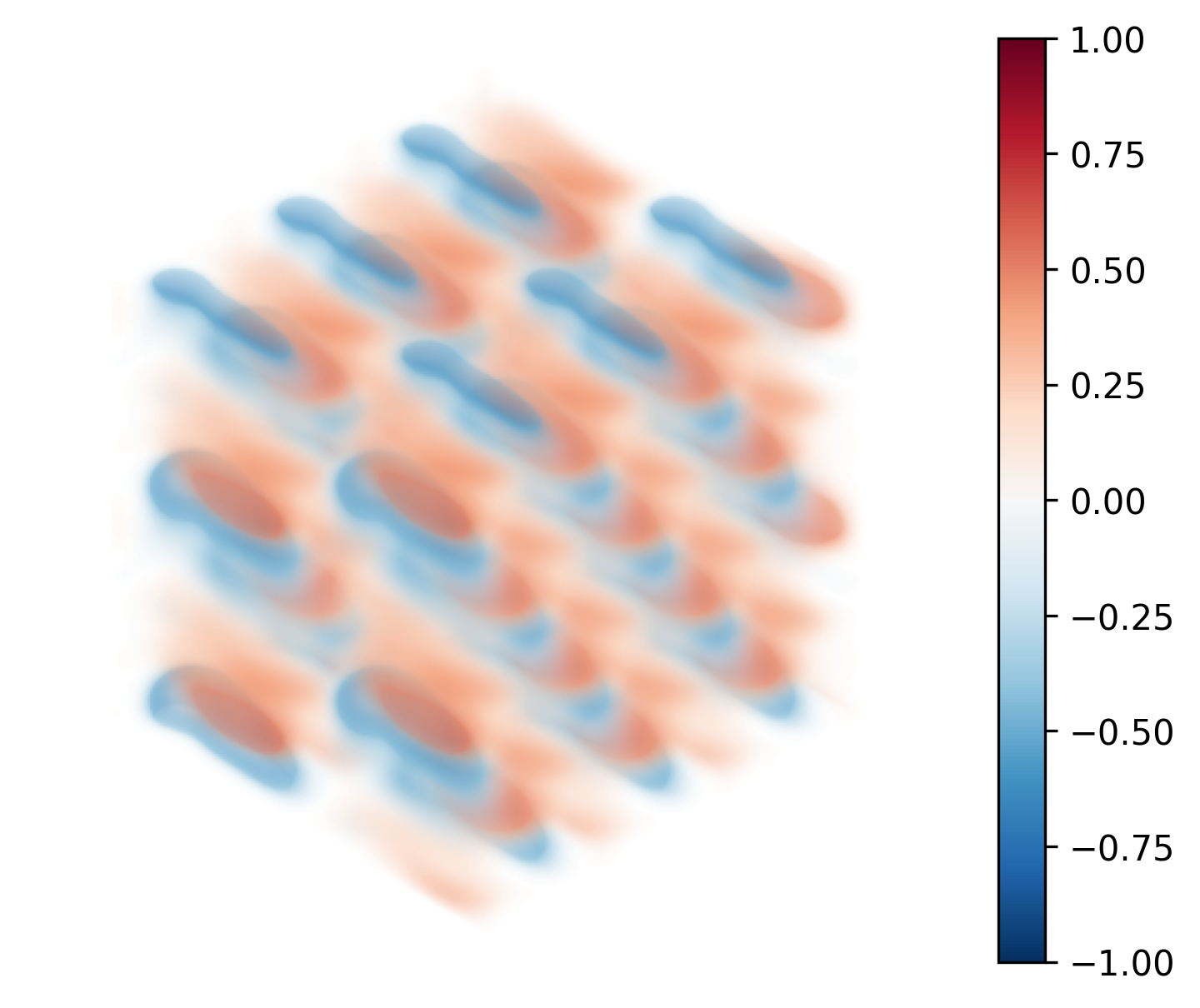}
        \label{}
     \end{subfigure}
    \begin{subfigure}[b]{0.32\textwidth}
         \centering
          \includegraphics[width=\textwidth]{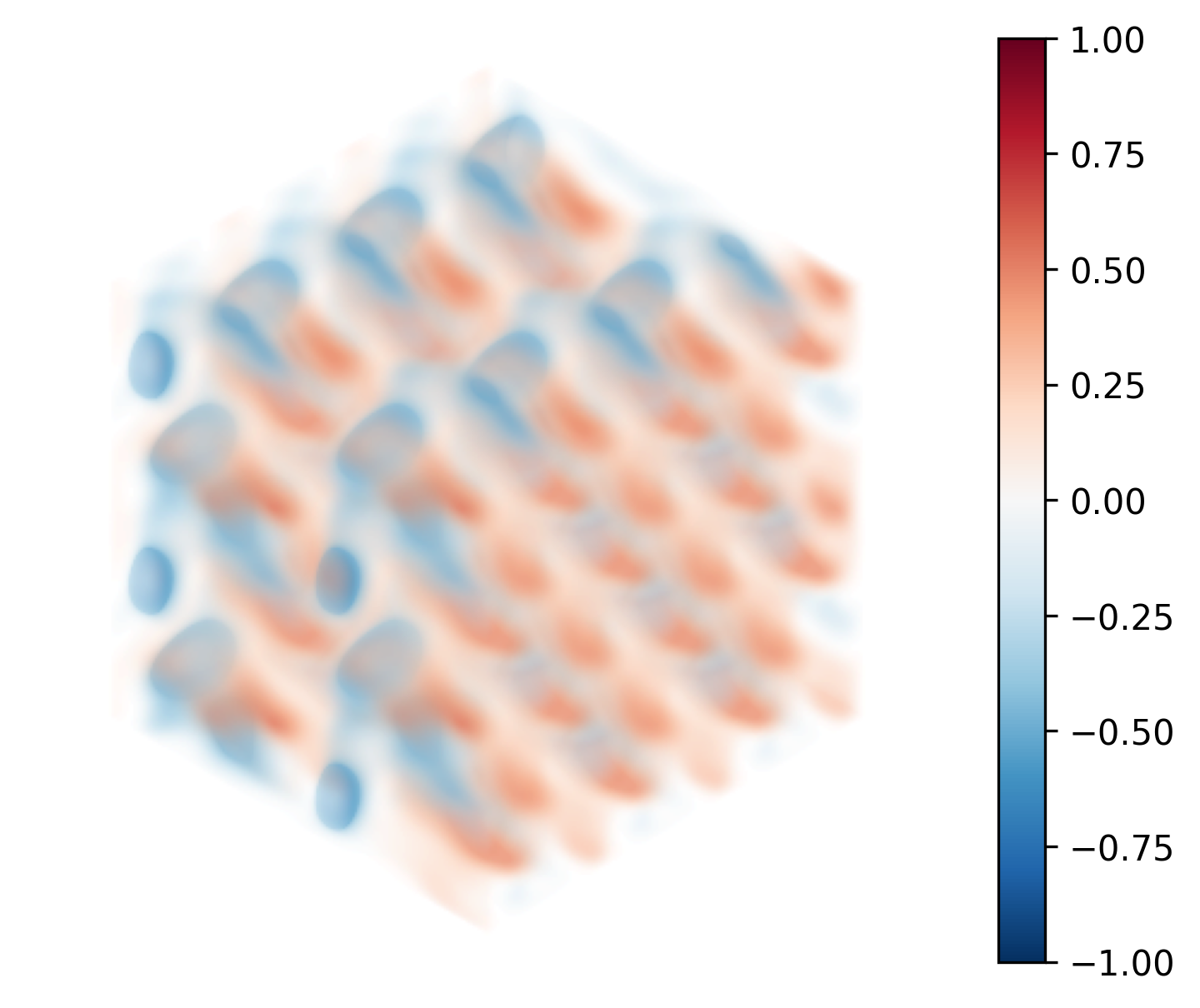}
         \label{}
     \end{subfigure}
         \begin{subfigure}[b]{0.32\textwidth}
         \centering
          \includegraphics[width=\textwidth]{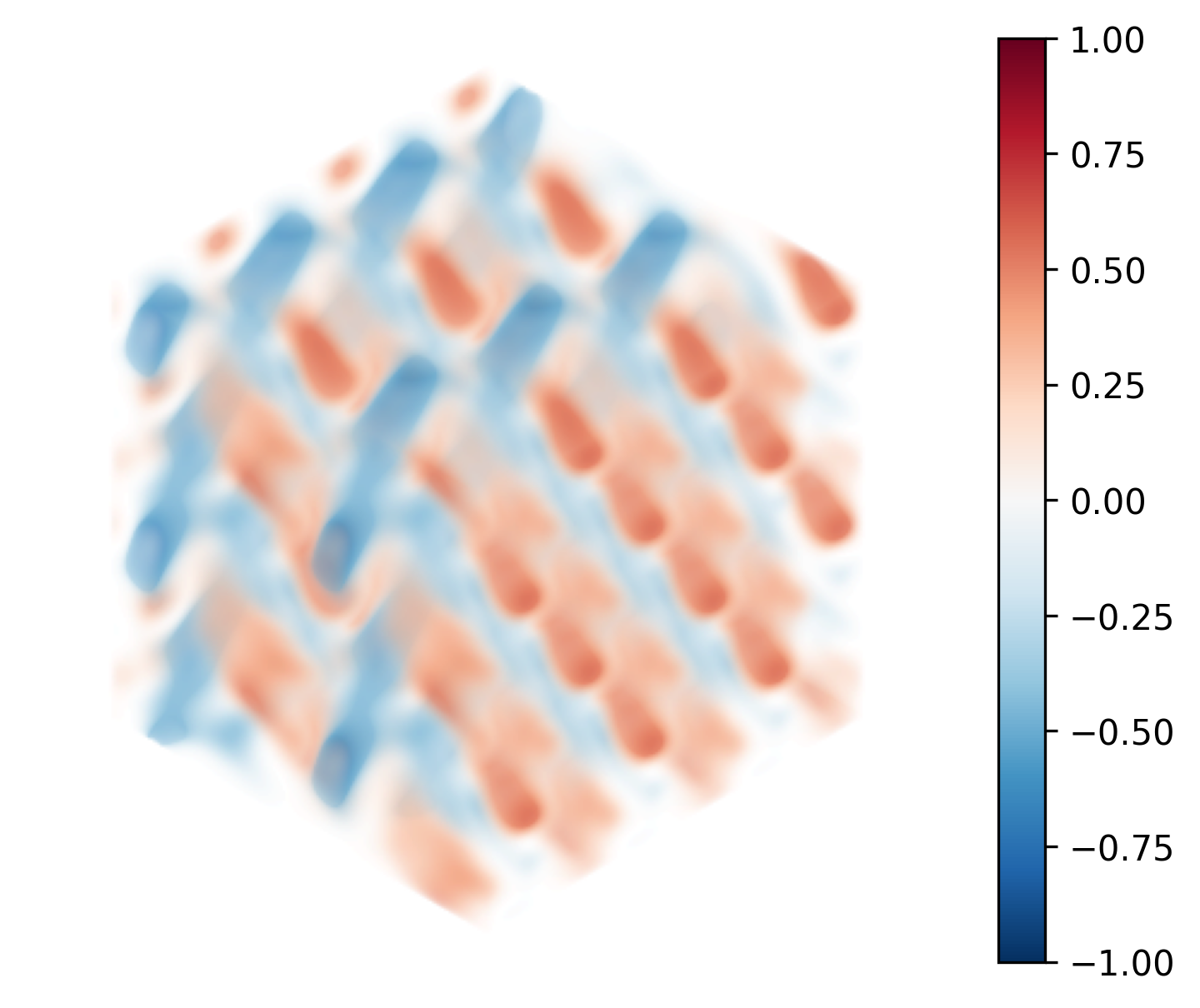}
         \label{}
     \end{subfigure}
        \caption{Examples of solution snapshots of the Korteweg-de Vries equation for the two different instances of the parameter and timesteps.}
        \label{fig:examples_solutions_kdv}
\end{figure}

As a third numerical example, we consider a 3-dimensional Korteweg-de Vries (KdV) equation. The KdV equation is a nonlinear dispersive PDE describing a wide range of physical systems, including internal ocean waves, plasma physics, lattice dynamics, and nonlinear optics \cite{korteweg1895xli}. The KdV can be written as follows:
\begin{equation}
\frac{\partial x(\bm{y}, t)}{\partial t} + \frac{1}{2}\Big(\frac{\partial x(\bm{y}, t)^2}{\partial y_1} + \frac{\partial x(\bm{y}, t)^2}{\partial y_2} + \frac{\partial x(\bm{y}, t)^2}{\partial y_3} \Big) +\Big(\frac{\partial^3 x(\bm{y}, t)}{\partial y_1^3} + \frac{\partial^3 x(\bm{y}, t)^2}{\partial y_2^3} + \frac{\partial^3 x(\bm{y}, t)^2}{\partial y_3^3} \Big) = -\mu f(x(\bm{y},t))\,,
\label{eq:kdv_3d}
\end{equation}
where $x(\bm{y},t)=x(y_1, y_2, y_3)$ is a scalar field defined on a three-dimensional periodic domain, $\mu$ is the hyper-diffusivity parameter of interest and
\begin{equation}
    f(x(\bm{y},t)) = \frac{\partial^4 x(\bm{y}, t)}{\partial y_1^4}+\frac{\partial^4 x(\bm{y}, t)}{\partial y_2^4}+\frac{\partial^4 x(\bm{y}, t)}{\partial y_3^4}\,.
\end{equation}
We obtain the FOM snapshots by solving \eqref{eq:kdv_3d} for different values of the parameter $\mu \in [0.0005, 0.05]$ for $40$ time steps with $\Delta t=0.05$s using \texttt{Exponax} \cite{koehler2024apebench}. \texttt{Exponax} employs a Fourier pseudo-spectral discretization in space combined with an exponential time-differencing Runge-Kutta scheme for temporal integration. The 3-dimensional spatial domain is discretized in $N=100$ bins per dimension, resulting in a one million-dimensional snapshots. We compute 100 FOM solutions in total, which are distributed into disjoint sets for training, validation and testing. The training set $X_{\text{train}}$ is composed of the 80 FOM solutions. Further, we keep 10 of the 100 FOM solutions for the validation set $X_{\text{valid}}$, while the test set $X_{\text{test}}$ is composed of the remaining 10 FOM solutions generated with 10 unseen parameter values.  Examples of test solutions can be found in Figure \ref{fig:examples_solutions_kdv}.

Similarly to the previous numerical example, we first apply POD to the snapshots to reduce the dimensionality from one million to $r=512$, posing a lower bound on the reconstruction error of $1.92 e^{-7}$. Again, it is worth noting that with the POD pre-processing step we simply want to decrease the snapshot dimensionality to be able to use light-weight AEs. We compare our inv-AEs with the convolutional AE introduced in \cite{fresca2021comprehensive} and a fully connected AE. We train the different AE models for 1000 epochs with AdamW optimizer and manifold dimensions $\in \{1, 5, 10,  50, 100\}$. For all the models, we use a learning rate of  $1e^{-4}$, weight decay equal to  $1e^{-4}$, and we employ early stopping if the validation loss does not improve for $500$ consecutive epochs.

\begin{figure}[h!]
    \centering
    \includegraphics[width=0.7\linewidth]{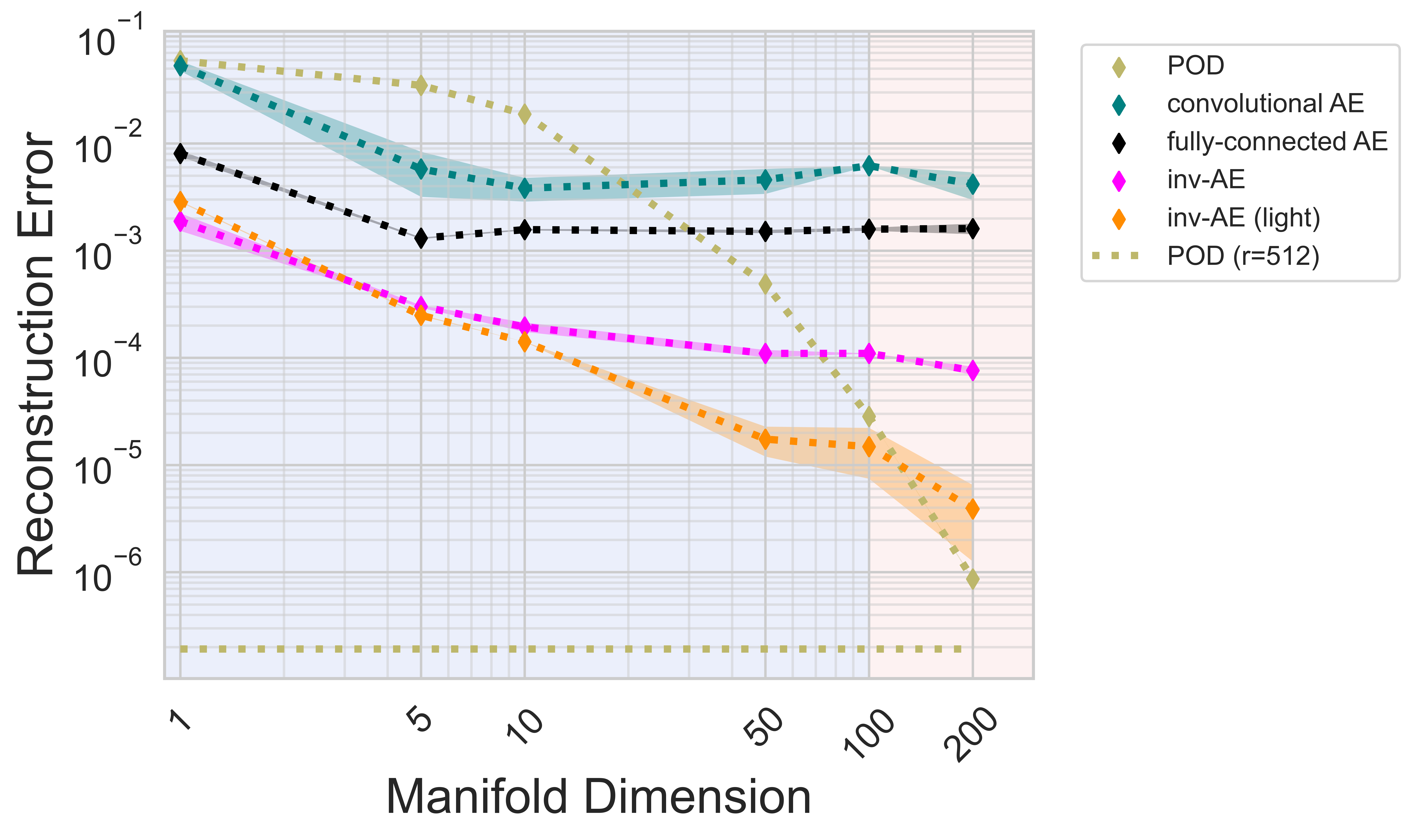}
    \caption{Reconstruction errors \eqref{eq:proj_error} for different dimensions of the trial manifold. The dashed line represents the mean and the shaded area the standard deviation across two different random seeds.}
    \label{fig:kdv_projection_error}
\end{figure}

In Figure \ref{fig:kdv_projection_error}, we show the reconstruction error \eqref{eq:proj_error} obtained using the convolutional AE, the fully connected AE, the inv-AEs, and the POD, and in Table \ref{tab:projection_errors_kdv} we report the best values. In addition, we report the reconstruction error when truncating the POD to $r=512$. Also in this case, both inv-AE variants reduce the reconstruction error compared to the convolutional AE and fully connected AE. Moreover, for both of the inv-AE, we observe an almost monotonic decay of the reconstruction error, which is neither the case for the convolutional nor the fully connected AE.  

\begin{figure}[h!]
    \centering
    \includegraphics[width=0.7\linewidth]{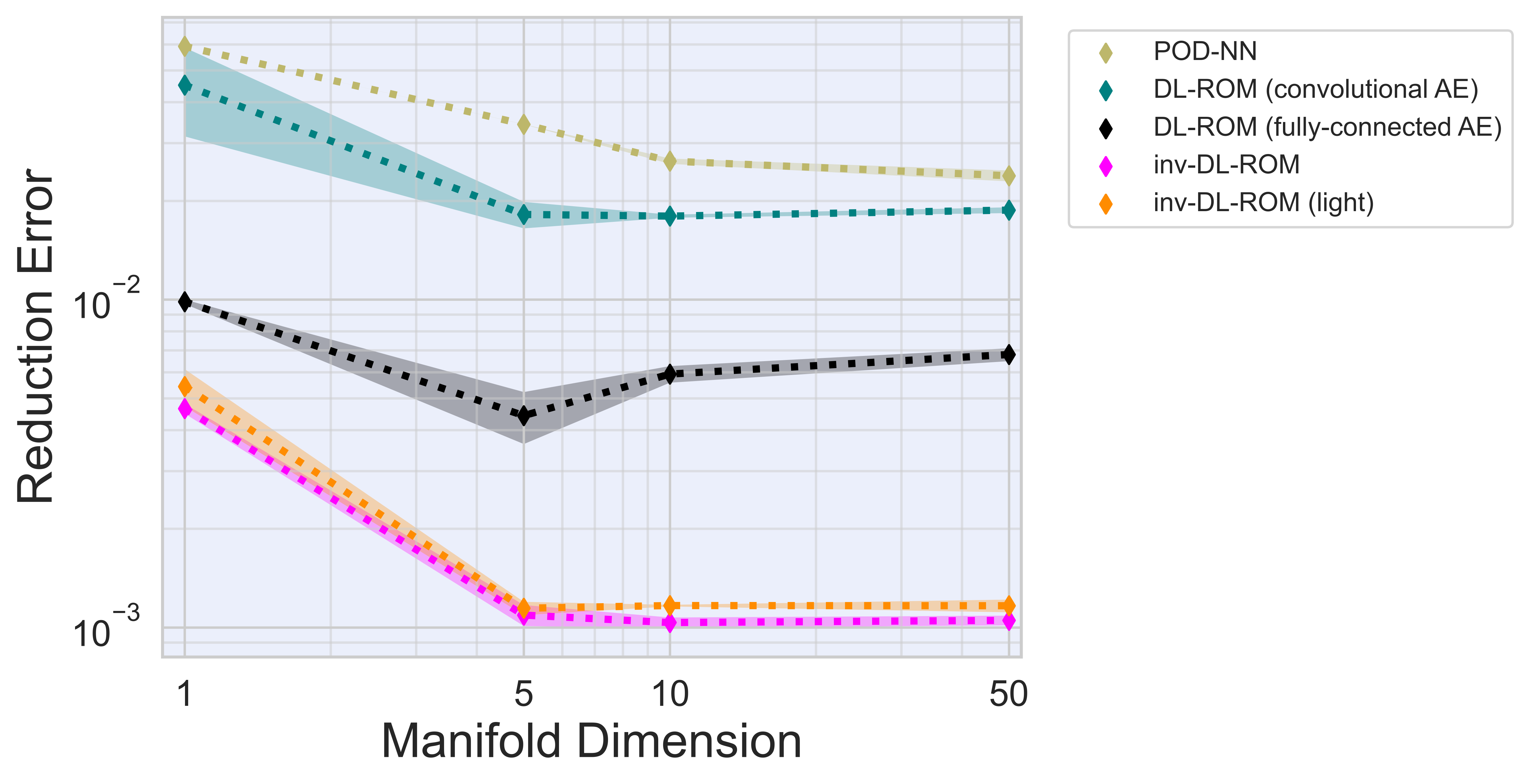}
    \caption{Reduction errors \eqref{eq:red_error} of DL-ROMs and inv-DL-ROMs. The dashed line represents the mean and the shaded area the standard deviation across two different random seeds.}
    \label{fig:kdv_reduction_error}
\end{figure}

Eventually, in Figure \ref{fig:kdv_reduction_error}, we show the reduction errors \eqref{eq:red_error} of POD-DL-ROMs and the POD-inv-DL-ROMs, while in Table \ref{tab:reduction_errors_kdv} we report the best values. Even in this case, by simply replacing the convolutional or fully connected AE with the inv-AEs, we are capable of significantly improving the reduction error of the POD-DL-ROM architecture.

\begin{figure*}[h!]
    \centering
    
    \begin{sideways} \makebox[0pt][l]{\hspace{-0.7cm} \shortstack[c]{{\bf \footnotesize FOM solution}}} \end{sideways}
    \begin{minipage}{0.95\linewidth}
    \centering
    \hspace{-0.12cm}
    \subfloat{\includegraphics[width=0.45\linewidth]{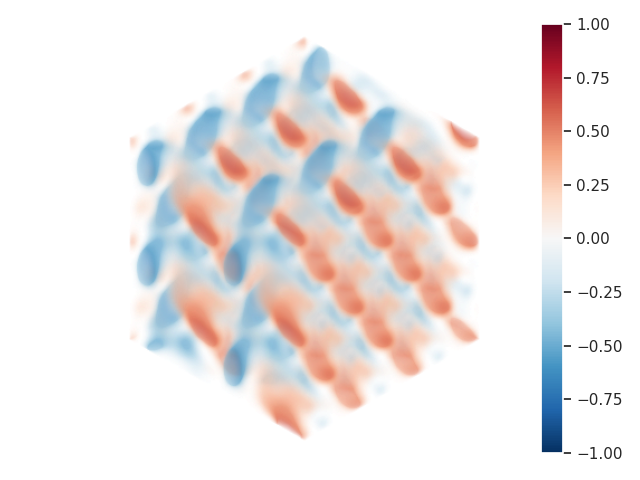}}
    \subfloat{\includegraphics[width=0.45\linewidth]{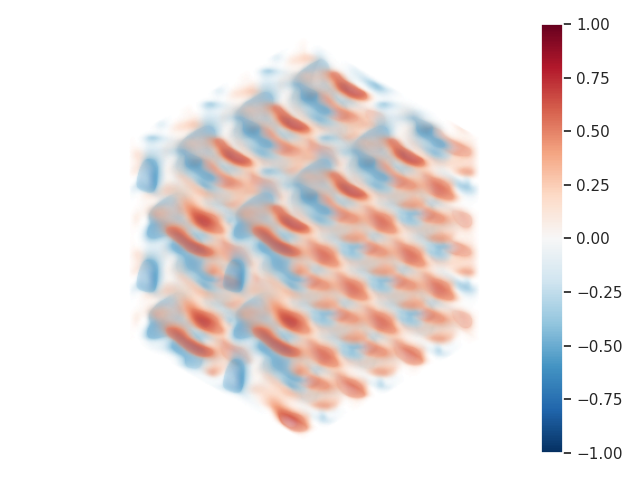}}
    \end{minipage}
    \hspace{-0.35cm}

    \begin{sideways} \makebox[0pt][l]{\hspace{-1.1cm} \shortstack[c]{{\bf \footnotesize Manifold dimension = 1}}} \end{sideways}
    \begin{minipage}{0.95\linewidth}
    \centering
    \hspace{-0.12cm}
    \subfloat{\includegraphics[width=0.45\linewidth]{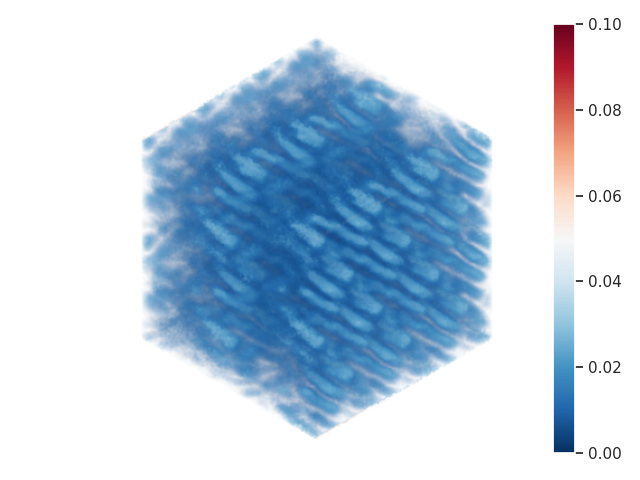}}
    \subfloat{\includegraphics[width=0.45\linewidth]{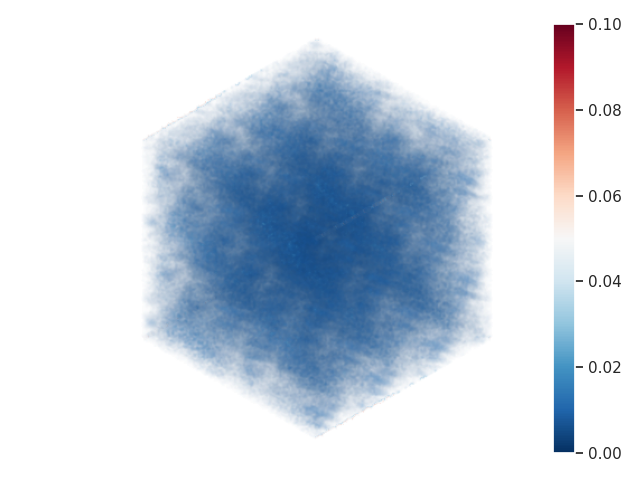}}
    \end{minipage}
    \hspace{-0.35cm}

    \begin{sideways} \makebox[0pt][l]{\hspace{-1.1cm} \shortstack[c]{{\bf \footnotesize Manifold dimension = 10}}} \end{sideways}
    \begin{minipage}{0.95\linewidth}
    \centering
    \hspace{-0.12cm}
    \subfloat{\includegraphics[width=0.45\linewidth]{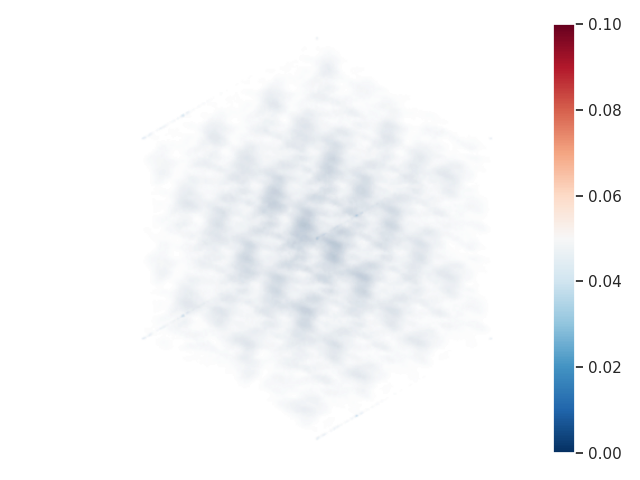}}
    \subfloat{\includegraphics[width=0.45\linewidth]{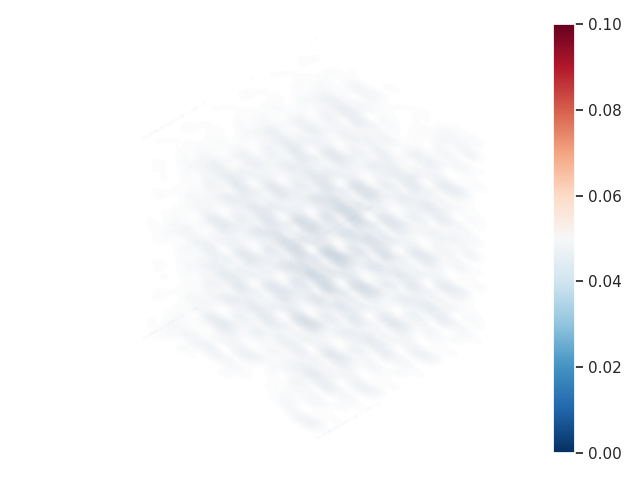}}
    \end{minipage}
    \hspace{-0.35cm}

        \begin{sideways} \makebox[0pt][l]{\hspace{-1.1cm} \shortstack[c]{{\bf \footnotesize Manifold dimension = 100}}} \end{sideways}
    \begin{minipage}{0.95\linewidth}
    \centering
    \hspace{-0.12cm}
    \subfloat{\includegraphics[width=0.45\linewidth]{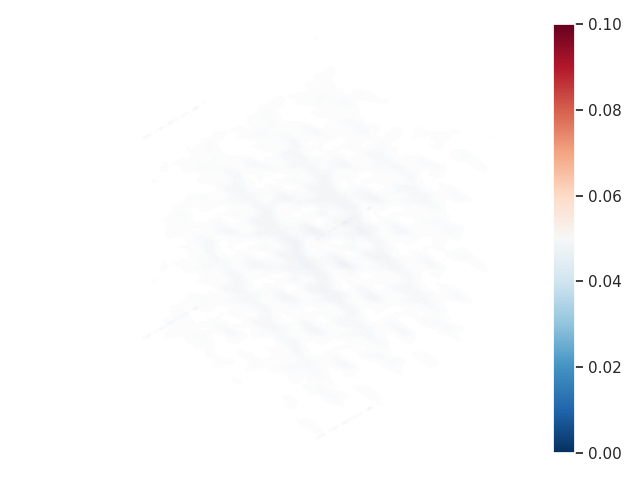}}
    \subfloat{\includegraphics[width=0.45\linewidth]{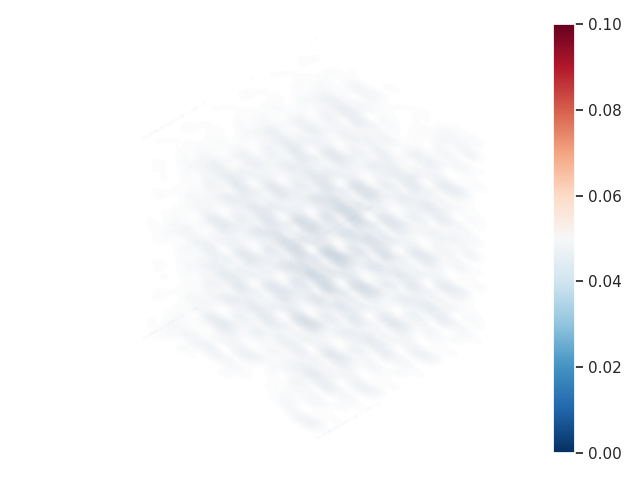}}
    \end{minipage}
    \hspace{-0.35cm}
    
    \vspace{-0.3cm}
    \caption{Qualitative comparison of the reconstructions of the inv-AE for different manifold dimensions. The first row shows two FOM solutions from the test set, while the remaining rows the absolute reconstruction error achieved by the different inv-AEs.}
    \label{fig:kdv_solutions}
\end{figure*}

Eventually, in Figure \ref{fig:kdv_solutions} we show the reconstruction errors of the inv-AE for increasing manifold dimensions, where we highlight the decrease trend of the absolute error.

\section{Conclusion}\label{sec:conclusion}

In this work, we propose an invertible AE architecture, named inv-AE, that allows for solving challenging dimensionality reduction problems. Inv-AE relies on {\em (i)} an invertible neural network to effectively learn an analytically invertible encoder-decoder scheme, on {\em (ii)} affine coupling layers to ensure numerical stability, and {\em (iii)} parameter sharing between encoder and decoder to reduce the number of learnable parameters. Moreover, we propose an integration of the inv-AE with popular data-driven MOR strategies such as DL-ROM and POD-DL-ROM, that we named inv-DL-ROM and POD-inv-DL-ROM.

In numerical experiments dealing with two challenging classes of high-dimensional problems, namely the Burgers' equation, the flow around an obstacle, and a Korteweg-de Vries equation, we show that inv-AE is capable of achieving a lower reconstruction error than traditional convolutional and fully connected AEs, and that its use in inv-DL-ROM and POD-inv-DL-ROM is beneficial for improving the reduction error of DL-ROM and POD-DL-ROM. Throughout our results, we show that inv-AE can mitigate the reconstruction-error plateau of traditional AEs architectures and the quality of its reconstruction gradually improves with the increment of the manifold dimension. However, the reduction error does not necessarily improves in the same way. This limitation might be intrinsic to the DL-ROM architecture and improving upon is an interesting future research direction.

While the inv-AE is analytically invertible by construction, the choice of the affine coupling layers is critical for its numerical invertibility. We experience that our affine coupling layers with clamping are not prone to training instabilities  and the use of spectral normalization (see \ref{sec:spectral_normalization}) appears to be only mildly beneficial. Future research will investigate these properties of the inv-AE to derive theoretical results and bounds for the proposed architectures. 

The current Inv-AE architecture relies on fully connected neural networks as the basic building blocks of the affine coupling transformations. Fully connected layers do not explicitly exploit any spatial structure of the (discretized) physical domain. This limitation may become particularly relevant when considering problems characterized by complex geometries, unstructured meshes, or varying spatial resolutions. In such settings, extending the reversible coupling blocks to incorporate more advanced architectures, such as graph neural networks, represents a promising research direction toimprove the scalability and generalization capabilities of inv-AE for large-scale problems.

\section*{Acknowledgements}
NB acknowledges the Project “Reduced Order Modeling and Deep Learning for the real- time approximation of PDEs (DREAM)” (Starting Grant No. FIS00003154), funded by the Italian Science Fund (FIS) - Ministero dell'Università e della Ricerca. 

\bibliographystyle{unsrt}  
\bibliography{references} 

\appendix

\section{Additional Results}
\subsection{Burgers Equation}

\begin{table}[h!]
\centering
\begin{tabular}{c|c|c|c|c|c}
\shortstack{\textbf{Manifold}\\ \textbf{Dimension}} & \shortstack{\textbf{POD}\\\textbf{}} & \shortstack{\textbf{Convolutional AE}\\ \textbf{}} & \shortstack{\textbf{Fully-connected AE}\\ \textbf{}} & \shortstack{\textbf{Inv-AE}\\ \textbf{}} & \shortstack{\textbf{Inv-AE }\\ \textbf{(light)}} \\ 
\hline 
1  & \(3.82{e}^{-2}\) & $9.13 e^{-3}$ & $\bm{8.71e^{-3}}$ & $8.91e^{-3}$ & $8.89e^{-3}$ \\
2  & \(2.70{e}^{-2}\) & $5.38e^{-3}$ & $4.57e^{-3}$& $4.27e^{-3}$ & $\bm{3.77e^{-3}}$ \\
3  & \(2.09{e}^{-2}\) & \(4.19e^{-4}\) & $1.94e^{-4}$ & $\bm{1.42e^{-4}}$ & $1.83e^{-4}$ \\
4  & \(1.73{e}^{-2}\) & \(3.36e^{-4}\) & $1.61e^{-4}$ & $\bm{1.20e^{-4}}$ & $1.49e^{-4}$ \\
5  & \(1.48{e}^{-2}\) & \(3.20e^{-4}\) & $1.49e^{-4}$ & $\bm{1.00e^{-4}}$ & $1.39e^{-4}$ \\
10  & \(8.60{e}^{-3}\) & \(2.77e^{-4}\) & $1.42e^{-4}$ & $\bm{8.08e^{-5}}$ & $1.28e^{-4}$ \\
20  & \(4.06{e}^{-3}\) & \(2.80e^{-4}\) & $1.39e^{-4}$ & $\bm{7.45e^{-5}}$ & $1.36e^{-4}$ \\
50  & \(6.46{e}^{-4}\) & \(3.31e^{-4}\) & $1.50e^{-4}$ & $\bm{6.61e^{-5}}$ & $8.92e^{-5}$ \\
100  & $\bm{3.46{e}^{-5}}$ & \(3.41e^{-4}\) & $1.33e^{-4}$ & $5.45e^{-5}$ & $6.96e^{-5}$ \\
150 & $\bm{1.94{e}^{-6}}$ & \(2.91e^{-4}\) & $1.62e^{-4}$ & $4.94e^{-5}$ & $5.87e^{-5}$ \\
200 & $\bm{1.18{e}^{-7}}$ & \(3.15e^{-4}\) & $1.56e^{-4}$ & $4.65e^{-5}$ & $3.40e^{-5}$ \\
\end{tabular}
\caption{Best reconstruction errors calculated on the test set achieved by the different approaches. We use \textbf{bold} to highlight the best values.}
\label{tab:projection_errors_burgers}
\end{table}
\begin{table}[h!]
\centering
\begin{tabular}{c|c|c|c|c|c}
\shortstack{\textbf{Manifold}\\ \textbf{Dimension}} & \shortstack{\textbf{POD-NN}\\ \textbf{}} &\shortstack{\textbf{DL-ROM}\\\textbf{(Conv. AE)}} & \shortstack{\textbf{DL-ROM}\\\textbf{(Fully-conn. AE)}} &\shortstack{\textbf{Inv-DL-ROM}\\\textbf{}} & \shortstack{\textbf{Inv-DL-ROM}\\\textbf{(light)}} \\
\hline
1 & - & $9.70e^{-3}$ & $\bm{9.60e^{-3}}$ & $9.67e^{-3}$ & $9.66e^{-3}$ \\
2 & $3.82e^{-2}$ & $6.78e^{-3}$ & $6.61e^{-3}$ & $3.28e^{-3}$ & $\bm{2.82e^{-3}}$ \\
3 & - & $1.29e^{-3}$ & $6.71e^{-4}$ & $4.27e^{-4}$ & $\bm{4.18e^{-4}}$ \\
4 & $2.71e^{-2}$ & $1.44e^{-3}$ & $6.49e^{-4}$ & $4.12e^{-4}$ & $\bm{3.91e^{-4}}$ \\
5 & - & $1.21e^{-3}$ & $6.33e^{-4}$ & $\bm{3.78e^{-4}}$ & $3.93e^{-4}$ \\
10 & $1.48e^{-2}$ & $1.40e^{-3}$ & $5.52e^{-4}$ & $\bm{3.76e^{-4}}$ & $3.83e^{-4}$ \\
20 & $8.63e^{-3}$ & $1.22e^{-3}$ & $6.06e^{-4}$ & $\bm{4.37e^{-4}}$ & $4.44e^{-4}$ \\
\end{tabular}
\caption{Best reduction errors calculated on the test set achieved by the different approaches. We use \textbf{bold} to highlight the best values.}
\label{tab:reduction_errors_burgers}
\end{table}

\subsection{Flow Around an Obstacle}

\begin{table}[h!]
\centering
\begin{tabular}{c|c|c|c|c|c}
\shortstack{\textbf{Manifold}\\ \textbf{Dimension}} & \shortstack{\textbf{POD}\\\textbf{}} & \shortstack{\textbf{Convolutional AE}\\ \textbf{}} & \shortstack{\textbf{Fully-connected AE}\\ \textbf{}} & \shortstack{\textbf{Inv-AE}\\ \textbf{}} & \shortstack{\textbf{Inv-AE }\\ \textbf{(light)}} \\ 
\hline 
1  & -  & $3.31e^{-2}$ & $\bm{1.94e^{-2}}$ & $2.16e^{-2}$ & $2.21e^{-2}$ \\
2  & $3.31e^{-2}$ & $1.53e^{-2}$ & $1.25e^{-2}$ & $1.23e^{-2}$ & $\bm{1.19e^{-2}}$ \\
3  & - & $1.16e^{-2}$ & $\bm{9.36e^{-3}}$ & $9.46e^{-3}$ & $9.66e^{-3}$ \\
4  & $1.30e^{-2}$ & $9.17e^{-3}$ & $8.52e^{-3}$ &$\bm{7.74e^{-3}}$ & $7.85e^{-3}$ \\
5  & - & $9.38e^{-3}$ & $8.13e^{-3}$ & $\bm{6.48e^{-3}}$ & $6.50e^{-3}$ \\
10  & $9.12e^{-3}$ & $1.33e^{-2}$ & $6.37e^{-3}$ & $3.70e^{-3}$ & $\bm{3.13e^{-3}}$ \\
20  & $6.59e^{-3}$ & $1.66e^{-2}$ & $6.26e^{-3} $ & $2.25e^{-3}$ & $\bm{2.18e^{-3}}$ \\
50  &$3.63e^{-3}$& $1.46e^{-2}$ & $6.05e^{-3}$ & $1.63e^{-3}$ & $\bm{1.58e^{-3}}$ \\
100  & $1.99e^{-3}$ & $1.39e^{-2}$ & $6.35e^{-3}$ & $1.18e^{-3}$ & $\bm{9.70e^{-4}}$ \\
150 & - & $9.63e^{-3}$ & $6.23e^{-3}$ & $9.16e^{-4}$  & $\bm{7.93e^{-4}}$  \\
200 & $9.36e^{-4}$ & $1.47e^{-2}$ & $6.44e^{-3}$ & $9.25e^{-4}$ & $\bm{7.06e^{-4}}$ \\
512 & $3.09e^{-4}$ & - & - & - & - \\
\end{tabular}
\caption{Best reconstruction errors calculated on the test set achieved by the different approaches. We use \textbf{bold} to highlight the best values.}
\label{tab:projection_errors_flow}
\end{table}
\begin{table}[h!]
\centering
\begin{tabular}{c|c|c|c|c|c}
\shortstack{\textbf{Manifold}\\ \textbf{Dimension}} & \shortstack{\textbf{POD-NN}\\ \textbf{}} &\shortstack{\textbf{DL-ROM}\\\textbf{(Conv. AE)}} & \shortstack{\textbf{DL-ROM}\\\textbf{(Fully-conn. AE)}} &\shortstack{\textbf{Inv-DL-ROM}\\\textbf{}} & \shortstack{\textbf{Inv-DL-ROM}\\\textbf{(light)}} \\
\hline
1 & - & $\bm{2.40e^{-2}}$             & $2.58e^{-2}$ & $3.23e^{-2}$ & $3.41e^{-2}$ \\
2 & $3.34e^{-2}$ & $1.30e^{-2}$  & $\bm{1.30e^{-2}}$ & $1.37e^{-2}$ & $1.36e^{-2}$ \\
3 & - & $1.02e^{-2}$             & $9.70e^{-3}$ & $\bm{9.61e^{-3}}$ & $9.65e^{-3}$ \\
4 & $1.32e^{-2}$ & $9.78e^{-3}$  & $9.29e^{-3}$ & $\bm{8.72e^{-3}}$ & $9.12e^{-3}$ \\
5 & - & $9.53e^{-3}$             & $9.29e^{-3}$ & $\bm{8.20e^{-3}}$ & $9.14e^{-3}$ \\
10 & $1.09e^{-2}$ & $9.12e^{-3}$ & $8.77e^{-3}$ & $\bm{8.22e^{-3}}$ & $8.56e^{-3}$ \\
20 & $1.00e^{-2}$ & $8.44e^{-3}$ & $8.92e^{-3}$ & $7.69e^{-3}$ & $\bm{7.63e^{-3}}$ \\
50 & $9.08e^{-3} $& $8.71e^{-3}$ & $8.63e^{-3}$ & $7.70e^{-3}$ & $\bm{7.17e^{-3}}$ \\
\end{tabular}
\caption{Best reduction errors calculated on the test set achieved by the different approaches. We use \textbf{bold} to highlight the best values.}
\label{tab:reduction_errors_flow}
\end{table}

\newpage
\subsection{Korteweg-de Vries}
\begin{table}[h!]
\centering
\begin{tabular}{c|c|c|c|c|c}
\shortstack{\textbf{Manifold}\\ \textbf{Dimension}} & \shortstack{\textbf{POD}\\\textbf{}} & \shortstack{\textbf{Convolutional AE}\\ \textbf{}} & \shortstack{\textbf{Fully-connected AE}\\ \textbf{}} & \shortstack{\textbf{Inv-AE}\\ \textbf{}} & \shortstack{\textbf{Inv-AE }\\ \textbf{(light)}} \\ 
\hline 
1  & $5.92e^{-2}$  & $4.75e^{-2}$ & $7.53e^{-3}$ & $\bm{1.53e^{-3}}$ & $2.83e^{-3}$ \\
5  & $3.49e^{-2}$ & $3.18e^{-3}$ & $1.29e^{-3}$ & $2.94e^{-4}$ & $\bm{2.46e^{-4}}$ \\
10  & $1.88e^{-2}$ & $2.88e^{-3}$ & $1.55e^{-3}$ & $1.75e^{-4}$ & $\bm{1.41e^{-4}}$ \\
50  & $4.89e^{-4}$ & $3.39e^{-3}$ & $1.46e^{-3}$ & $1.01e^{-4}$ & $\bm{1.20e^{-5}}$ \\
100  & $2.84e^{-5}$ & $6.18e^{-3}$ & $1.53e^{-3}$ & $1.10e^{-4}$ & $\bm{7.45e^{-6}}$ \\
200 & $\bm{8.48e^{-7}}$ & $2.97e^{-3}$ & $1.42e^{-3}$ & $6.94e^{-5}$ & $1.26e^{-6}$ \\
512 & $1.92e^{-7}$ & - & - & - & - \\
\end{tabular}
\caption{Best reconstruction errors calculated on the test set achieved by the different approaches. We use \textbf{bold} to highlight the best values.}
\label{tab:projection_errors_kdv}
\end{table}

\begin{table}[h!]
\centering
\begin{tabular}{c|c|c|c|c|c}
\shortstack{\textbf{Manifold}\\ \textbf{Dimension}} & \shortstack{\textbf{POD-NN}\\ \textbf{}} &\shortstack{\textbf{DL-ROM}\\\textbf{(Conv. AE)}} & \shortstack{\textbf{DL-ROM}\\\textbf{(Fully-conn. AE)}} &\shortstack{\textbf{Inv-DL-ROM}\\\textbf{}} & \shortstack{\textbf{Inv-DL-ROM}\\\textbf{(light)}} \\
\hline
1 & $5.92e^{-2}$ & $3.15e^{-2}$  & $9.70e^{-3}$ & $4.49e^{-3}$ & $\bm{4.73e^{-3}}$ \\
5 & $3.42e^{-2} $& $1.65e^{-2}$  & $3.63e^{-3}$ & $\bm{1.01e^{-3}}$ & $1.09e^{-3}$ \\
10 & $2.59e^{-2}$ & $1.78e^{-2}$ & $5.58e^{-3}$ & $\bm{9.97e^{-4}}$ & $1.16e^{-3}$ \\
50 & $2.29e^{-2}$ & $1.84e^{-2}$ & $6.49e^{-3}$ & $\bm{1.02e^{-3}}$ & $1.11e^{-3}$ \\
\end{tabular}
\caption{Best reduction errors calculated on the test set achieved by the different approaches. We use \textbf{bold} to highlight the best values.}
\label{tab:reduction_errors_kdv}
\end{table}

\clearpage

\section{Neural Network Architectures}\label{sec:nn_architecture}

Code available at: \url{https://github.com/nicob15/Inv-AE-Invertible-Autoencoder-for-Dynamical-Systems}

\subsection{Invertible AE}

The inv-AE is composed of 5 invertible layers $L$. Each reversible layer is based on a multiplicative affine coupling transformation, where the input vector is split into two parts $\mathbf{x}=[\mathbf{x}_1,\mathbf{x}_2]$. The forward transformation is defined as
\begin{equation*}
    \begin{split}
        \bm{y}_1 &= \bm{x}_1 \odot \exp(\bm{s}_2(\bm{x}_2);\boldsymbol{\theta}_{\bm{s}_2}) + \bm{t}_2(\bm{x}_2;\boldsymbol{\theta}_{\bm{t}_2})\\
        \bm{y}_2 &= \bm{x}_2 \odot \exp(\bm{s}_1(\bm{y}_1);\boldsymbol{\theta}_{\bm{s}_1}) + \bm{t}_1(\bm{y}_1; \boldsymbol{\theta}_{\bm{t}_1}) \\
    \end{split}\, ,
\end{equation*}
while the inverse transformation can be analytically computed as
\begin{equation*}
    \begin{split}
        \bm{x}_1 &= (\bm{y}_1 - \bm{t}_2(\bm{x}_2;\boldsymbol{\theta}_{\bm{t}_2})) \odot \exp(-\bm{s}_2(\bm{x}_2;\boldsymbol{\theta}_{\bm{s}_2})) \, , \\
        \bm{x}_2 &= (\bm{y}_2 - \bm{t}_1(\bm{y}_1;\boldsymbol{\theta}_{\bm{t}_1})) \odot \exp(-\bm{s}_1(\bm{y}_1;\boldsymbol{\theta}_{\bm{s}_1}))\, . \\
    \end{split}
\end{equation*}
In the case of the inv-AE, the scale and translation operators $\bm{s}_1, \bm{s}_2, \bm{t}_1, \bm{t}_2$ are represented by fully connected neural networks $N/2 \rightarrow 512 \rightarrow N/2$ with GeLU activation. In the case of the inv-AE (light), instead, we use 3 fully connected layers $N/2 \rightarrow 128 \rightarrow 128 \rightarrow N/2$ with GeLU actuvation to compensate for the reduction of the paramenters. The resulting architecture combines nonlinear dimensionality reduction with an explicitly invertible latent representation, enabling the learning of compact reduced-order coordinates while preserving a stable mapping between the latent and full-order spaces.

\subsection{1-dimensional Convolutional AE}
For one-dimensional spatial fields, namely the 1-dimensional Burgers' equation, we employ a convolutional autoencoder composed of a four-layer convolutional encoder and a transposed-convolutional decoder. The same architecture was used in \cite{lee2020model}.  The encoder is composed of 4 1-dimensional convolutional layers with ELU activations:
\begin{equation}
\begin{split}
&\text{Conv1:} \quad C \rightarrow 8 \qquad k=25, s=2, p=1, \\
&\text{Conv2:} \quad 8 \rightarrow 16, \qquad k=25, s=2, p=1, \\
&\text{Conv3:} \quad 16 \rightarrow 32, \qquad k=25, s=1, p=1, \\
&\text{Conv4:} \quad 32 \rightarrow 64, \qquad k=25, s=2, p=1, \\
\end{split}
\end{equation}
where $C$ denotes the number of channels, $k$ the kernel size, $s$ the stride length, and $p$ the padding dimension. The feature maps produced by the final convolutional layer are flattened and projected onto a low-dimensional latent space through a fully connected layer.
The decoder mirrors the encoder architecture. The latent vector is first mapped through a fully connected layer and reshaped into a tensor with 64 channels, which is then processed by four transposed convolutional layers:
\begin{equation}
    \begin{split}
&\text{Deconv1:} \quad 64 \rightarrow 32, \qquad k=25, s=1, p=0, \\
&\text{Deconv2:} \quad 32 \rightarrow 16, \qquad k=25, s=1, p=1, \\
&\text{Deconv3:} \quad 16 \rightarrow 8, \qquad k=25, s=2, p=0, \\
&\text{Deconv4:} \quad 8 \rightarrow C, \qquad k=25, s=2, p=0, \\
\end{split}
\end{equation}
with ELU activations applied after each layer except the final output layer. The relatively large kernel size ($k=25$) allows the network to capture long-range spatial correlations, while the strided convolutions progressively reduce the spatial resolution and extract increasingly abstract features before projection onto the latent manifold.

\subsection{2-dimensional Convolutional AE}
For the 2-dimensional and 3-dimensional test cases, we employ the convolutional autoencoder introduced in \cite{fresca2022pod}. The encoder consists of 4 2-dimensional convolutional layers, each employing ($3\times3$) kernels and ELU activation functions. The architecture is given by
\begin{equation}
\begin{split}
&\text{Conv1:} \quad 2 \rightarrow 32,\qquad k=3, s=1,\\
&\text{Conv2:} \quad 32 \rightarrow 32,\qquad k=3,; s=1,\\
&\text{BatchNorm}, \\
&\text{Conv3:} \quad 32 \rightarrow 32,\qquad k=3, s=1,\\
&\text{Conv4:} \quad 32 \rightarrow 32,\qquad k=3, s=1,\\
&\text{BatchNorm}. \\
\end{split}
\end{equation}
No pooling or strided convolutions are employed. Instead, the dimensionality reduction is performed through the fully connected layers following the convolutional feature extractor. After the final convolutional layer, the feature maps are flattened and projected onto the chosen manifold of dimension $n$ through 2 fully connected layers.
The latent vector is first projected back to the convolutional feature space using a fully connected layer and then reconstructed through 4 transposed convolutional layers
\begin{equation}
\begin{split}
&\text{Deconv1:} \quad 32 \rightarrow 32,\qquad k=3, s=1,\\
&\text{Deconv2:} \quad 32 \rightarrow 32,\qquad k=3, s=1,\\
&\text{BatchNorm},\\
&\text{Deconv3:} \quad 32 \rightarrow 32,\qquad k=3, s=1,\\\
&\text{Deconv4:} \quad 32 \rightarrow 2,\qquad k=3, s=1, \\
\end{split}
\end{equation}
with ELU activations applied after all layers except the final output layer. The use of small ($3\times3$) kernels enables the network to efficiently capture local spatial correlations the underlying field. Moreover, by avoiding aggressive downsampling operations, the architecture retains fine-scale spatial information that is essential for accurately reconstructing high-dimensional dynamical states.

\subsection{Fully Connected AE}
The encoder is composed of five fully connected layers, mapping the input state from dimension $N$ to a lower-dimensional latent space of dimension $n$ through the sequence $N\rightarrow2048\rightarrow1024\rightarrow512\rightarrow256\rightarrow n$. The decoder expands the latent representation back to the original state dimension through the sequence $n\rightarrow256\rightarrow512\rightarrow1024\rightarrow2048\rightarrow N$. Leaky-ReLU activation functions are applied after each hidden layer.

\subsection{Fully Connected Reduced-order Model}
In the proposed implementation, the $\xi$ is composed of 3 fully connected layers with architecture $(\boldsymbol{\mu}, t)\rightarrow 256 \rightarrow 256 \rightarrow n$, where $p$ denoted the dimension of the parameter space. The first two layers are followed by Leaky-ReLU nonlinear activation functions, while the final layer directly outputs the latent variables. Unlike the encoder, which extracts latent coordinates from observed states, this network learns a direct approximation of the latent manifold as a function of the governing parameters and time.

\section{Spectral Normalization}\label{sec:spectral_normalization} 
The training stability of NNs is heavily affected by the Lipschitz constant of networks themselves. Therefore, controlling the Lipschitz constant and preventing its rapid growth is crucial from avoiding exploding gradients and for the loss convergence. Spectral normalization \cite{miyato2018spectral} is a type of weight normalization that can improve performance of the NN architectures suffering from training instabilities, such as generative adversarial networks \cite{goodfellow2020generative}.

Given a NN layer with weight matrix $W$, the spectral norm of $W$ equals the largest singular value $\sigma_{\max}(W)$, i.e., ${\sigma_{\max}(W)}= \|W \|_2$. Spectral normalization then uses the spectral norm of $W$ for normalization such that $\sigma(W)=1$, i.e.,
\begin{align*}
    W_{\text{SN}} = \frac{W}{\sigma_{\max}(W)}\, ,
\end{align*}
where $W_{\text{SN}}$ is the spectrally normalized weight matrix. 

If the activation functions have Lipschitz constant equal to $1$,\footnote{This is the case for many activation functions such as ReLU and leaky ReLU.} the Lipschitz constant of a NN can be bounded from about by the product of the spectral norms of the the weight matrices of each layer $l$
\begin{equation}
    L \leq \prod_{l=1}^{N_L} \sigma_{\max}(W^l)\, ,
\end{equation}
where $N_L$ indicates the number of layers of the network and $W^l$ the weight matrix associated with $l^{\text{th}}$ layer. Therefore, if all layers of the network are spectrally normalized, then the Lipschitz constant of the network is bounded from above by $1$.

\begin{figure}[h!]
\centering
    \begin{subfigure}[b]{0.49\textwidth}
        \centering
        \includegraphics[width=\textwidth]{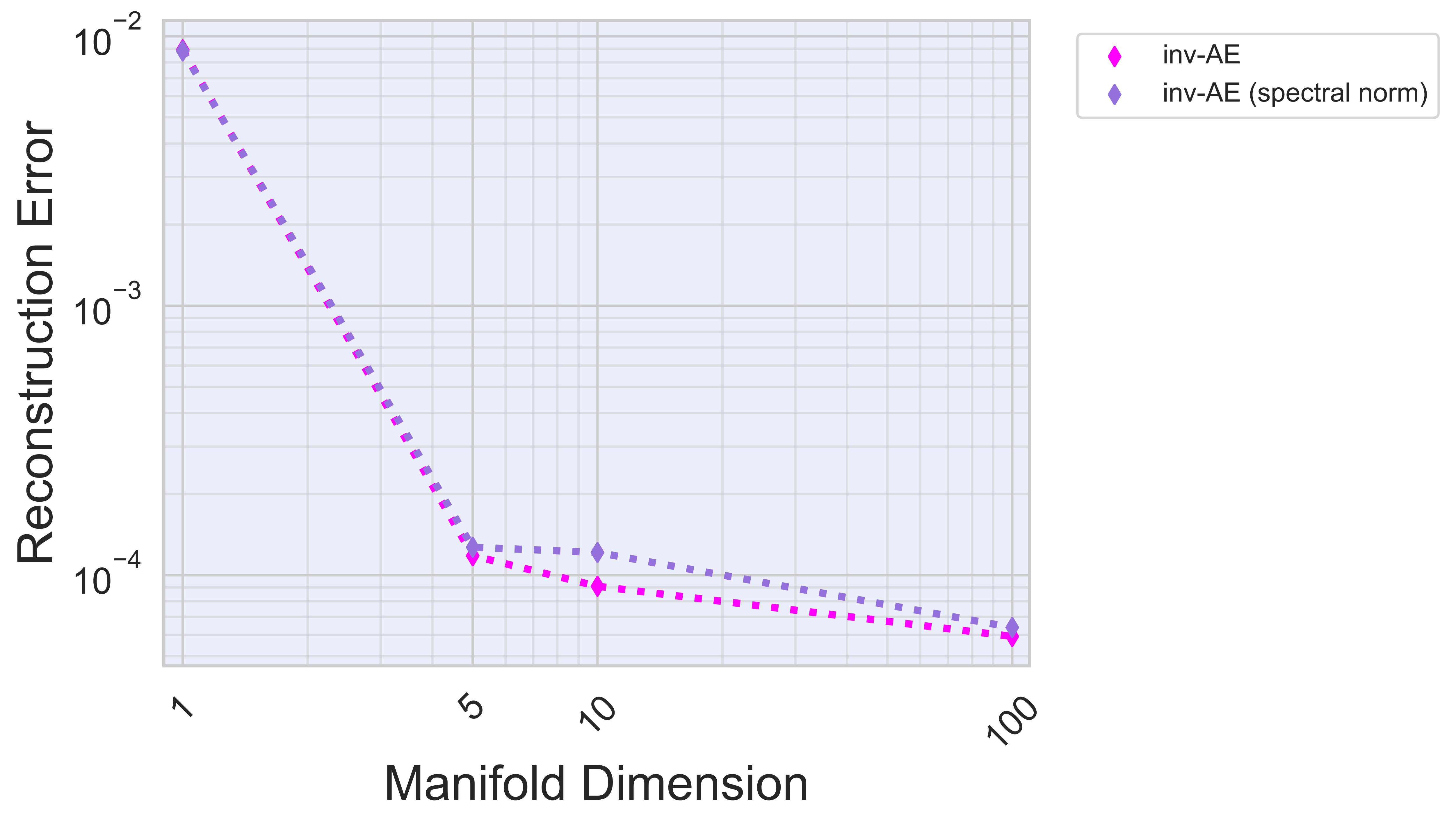}
        \label{fig:14a}
        \caption{Burgers' equation.}
     \end{subfigure}
    \begin{subfigure}[b]{0.49\textwidth}
         \centering
          \includegraphics[width=\textwidth]{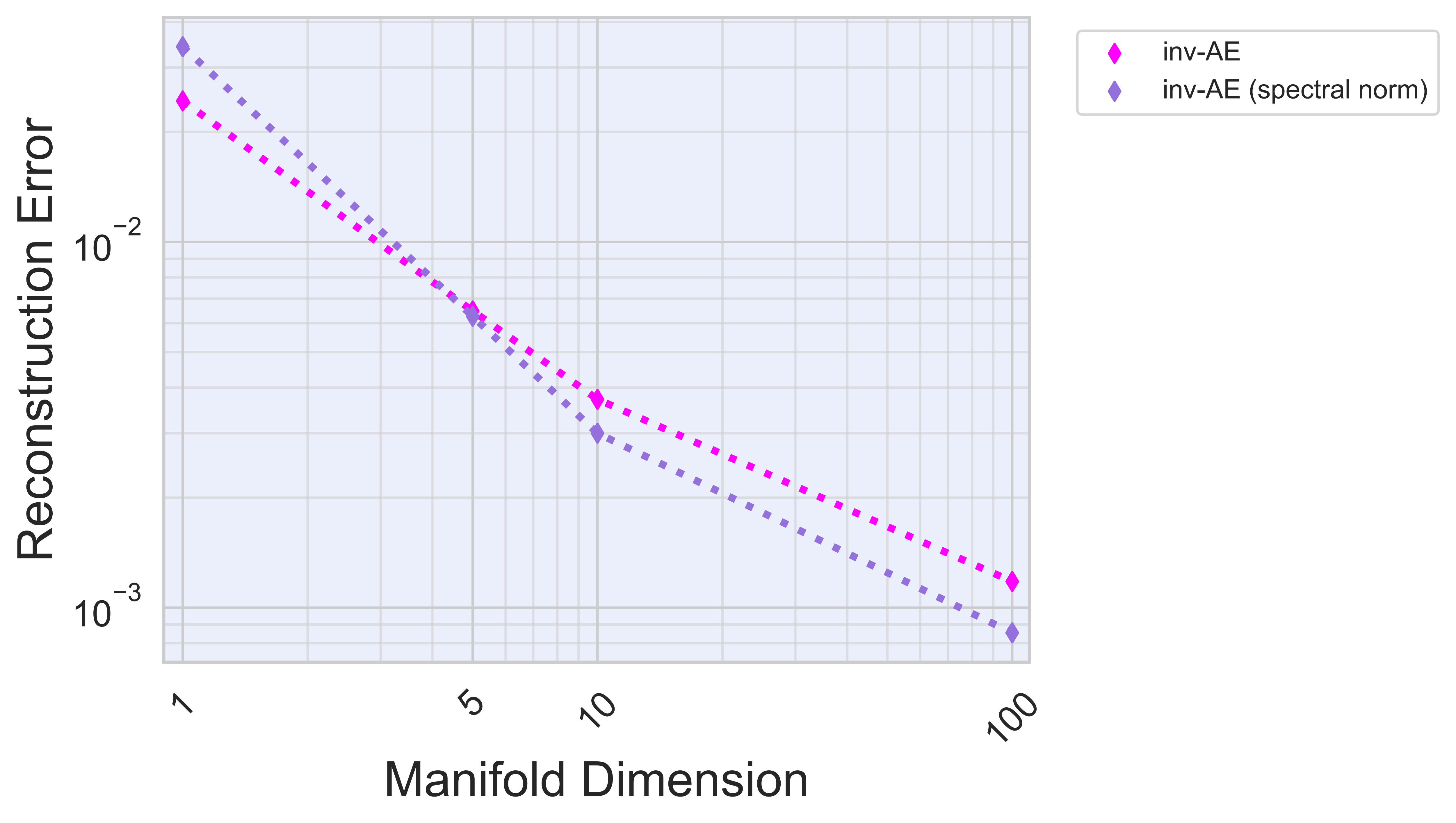}
         \label{fig:14b}
         \caption{Flow around obstacle.}
     \end{subfigure}
         \begin{subfigure}[b]{0.49\textwidth}
         \centering
          \includegraphics[width=\textwidth]{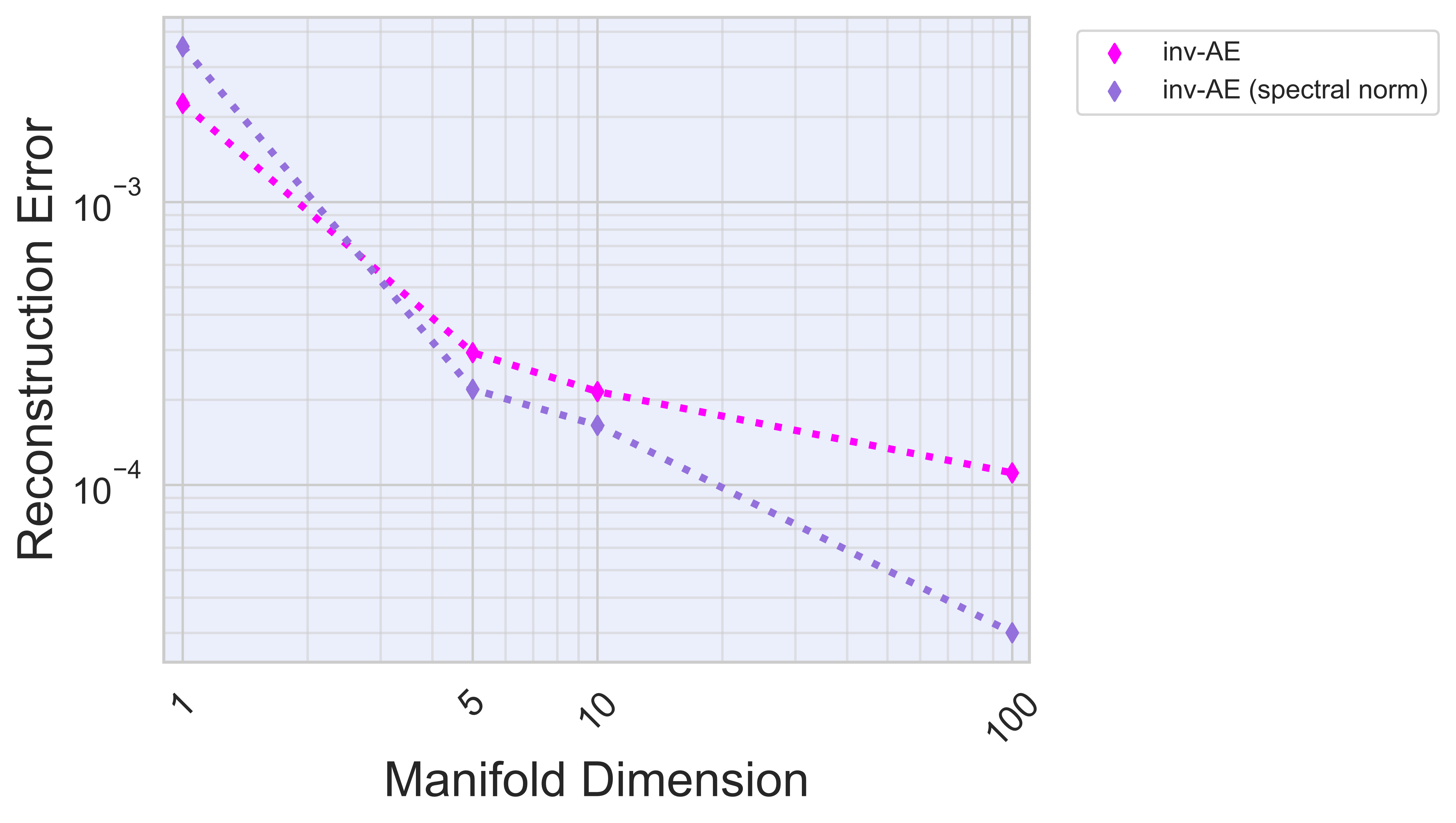}
         \label{fig:14c}
         \caption{Korteweg-de Vries.}
     \end{subfigure}
        \caption{Projection error with and without spectral normalization for the three different test cases.}
        \label{fig:spectral_norm_ablation}
\end{figure}

In Figure \ref{fig:spectral_norm_ablation} we show the effect of spectral normalization on the reconstruction errors of the AE for the three different test cases.

\end{document}